\documentclass[10pt,journal,compsoc]{IEEEtran}
\usepackage{color,soul}
\usepackage{graphics}
\usepackage{epsfig}
\usepackage{graphicx}
\usepackage{epstopdf}
\usepackage{amsmath}
\usepackage{paralist}
\usepackage[font=footnotesize]{caption}
\usepackage[font=footnotesize]{subcaption}
\usepackage{color,soul}
\usepackage[utf8]{inputenc}
\usepackage[T1]{fontenc}
\usepackage{lmodern}
\usepackage{booktabs}
\usepackage{verbatim}
\usepackage{multirow}
\usepackage{listings}
\usepackage[letterpaper, margin=0.75in]{geometry}
\usepackage{wrapfig}
\usepackage{paralist}
\usepackage{algorithm}
\usepackage{algpseudocode}
\usepackage[english]{babel}
\usepackage{blindtext}
\usepackage[colorlinks=true]{hyperref}
\usepackage[super]{nth}
\usepackage[switch]{lineno}
\usepackage{ulem}

\hypersetup{urlcolor=blue, citecolor=blue}

\algrenewcommand\algorithmicrequire{\textbf{Input:}}
\algrenewcommand\algorithmicensure{\textbf{Output:}}

\makeatletter
\def\BState{\State\hskip-\ALG@thistlm}
\makeatother

\begin{document}

\title{Extended Isolation Forest}
\author{Sahand~Hariri,~\IEEEmembership{}
	Matias~Carrasco~Kind,~\IEEEmembership{}
	Robert~J.~Brunner~\IEEEmembership{}
	\IEEEcompsocitemizethanks{
		\IEEEcompsocthanksitem S.Hariri is with the Department of Mechanical Science and Engineering, University of Illinois at Urbana-Champaign.\protect\\
		E-mail: sahandha@gmail.com

		\IEEEcompsocthanksitem M. Carrasco Kind is with The National Center for Supercomputing Applications at University of Illinois at Urbana-Champaign.\protect\\
		E-mail: mcarras2@illinois.edu

		\IEEEcompsocthanksitem R. J. Brunner is with The National Center for Supercomputing Applications at University of Illinois at Urbana-Champaign.\protect\\
		E-mail: bigdog@illinois.edu
	}

	\thanks{Manuscript received April 10, 2018}}


\IEEEtitleabstractindextext{%
	\begin{abstract}
		We present an extension to the model-free anomaly detection algorithm, Isolation Forest. This extension, named Extended Isolation Forest (EIF), resolves issues with assignment of anomaly score to given data points. We motivate the problem using heat maps for anomaly scores. These maps suffer from artifacts generated by the criteria for branching operation of the binary tree. We explain this problem in detail and demonstrate the mechanism by which it occurs visually. We then propose two different approaches for improving the situation. First we propose transforming the data randomly before creation of each tree, which results in averaging out the bias. Second, which is the preferred way, is to allow the slicing of the data to use hyperplanes with random slopes. This approach results in remedying the artifact seen in the anomaly score heat maps. We show that the robustness of the algorithm is much improved using this method by looking at the variance of scores of data points distributed along constant level sets. We report AUROC and AUPRC for our synthetic datasets, along with real-world benchmark datasets.  We find no appreciable difference in the rate of convergence nor in computation time between the standard Isolation Forest and EIF.
	\end{abstract}

	}

\maketitle

\section{Introduction}\label{sec:Intro}

In the age of big data and high volume information, anomaly detection finds many areas of application, including network security, financial data, medical data analysis, and discovery of celestial events from astronomical surveys, among many more. The need for reliable and efficient algorithms is plentiful, and there are many techniques that have been developed over the years to address this need including multivariate data \cite{Rocke1996} and more recently, streaming data with need to updates on data with missing variables \cite{pevny2016loda}. For a survey in research in anomaly detection see \cite{chandola2009anomaly, emmott2015meta}. Among the different anomaly detection algorithms, Isolation Forest \cite{liu2008isolation,Liu:2012:IAD:2133360.2133363} is one with unique capabilities. It is a model free algorithm that is computationally efficient, can easily be adapted for use with parallel computing paradigms \cite{hariri2018batch}, and has been proven to be very effective in detecting anomalies \cite{Susto2017}. The main advantage of the algorithm is that it does not rely on building a profile for data in an effort to find samples that do not conform to this profile. Rather, it utilizes the fact that anomalous data are ``few and different''. Most anomaly detection algorithms find anomalies by understanding the distribution of their properties and isolating them from the rest of normal data samples \cite{Chen2011,Noto2010,Das2016}. In an Isolation Forest, data is sub-sampled, and processed in a tree structure based on random cuts in the values of randomly selected features in the dataset. Those samples that travel deeper into the tree branches are less likely to be anomalous, while shorter branches  are indicative of anomaly. As such, the aggregated lengths of the tree branches provide for a measure of anomaly or an ``anomaly score'' for every given point. However, before using this anomaly score for processing of data, there are issues with the standard Isolation Forest algorithm that need to be addressed.

It turns out that while the algorithm is computationally efficient, it suffers from a bias that arises because of the way the branching of the trees takes place. In this paper we study this bias and discuss where it comes from and what effects it has on the anomaly score of a given data point. We further introduce an extension to the Isolation Forest, named Extended Isolation Forest (EIF), that remedies this shortcoming. While the basic idea is similar to the Isolation Forest, the details of implementation are modified to accommodate a more general approach.

As we will see later, the proposed extension relies on the use of hyperplanes with random slopes (non-axis-parallel) for splitting the data in creating the binary search trees. The idea of using hyperplanes with random slopes in the context of Isolation Forest with random choice of the slope is mentioned in \cite{liu2010detecting}. However, the main focus there is to find optimal slicing planes in oder to detect clustered anomalies. No further discussion on developing this method, or why it might benefit the algorithm is presented. The methods for choosing random slopes for hyperplanes, the effects of these choices on the algorithm, and the consequences of this, especially in higher dimensional data, are also omitted. Using oblique hyperplanes in tree based methods has also been studied in random forests \cite{menze2011oblique} but again, not in the context of Isolation Forest. An approach to anomaly detection which utilizes the concept of isolation is presented in \cite{bandaragoda2014efficient, bandaragoda2018isolation} where a nearest neighbor algorithm instead of a tree based method is used for isolating data points. While this algorithm performs better than the Isolation Forest by various metrics, and while the problem of the isolation forest discussed here is remedied because of the fundamentally different nature of achieving isolation, a discussion of what the causes of the problem with Isolation Forest is missing. Here we show one can achieve better performance by adjusting the algorithm and utilizing the tree based nature of the algorithm without having to resort to fundamentally different algorithms such as nearest neighbor.

As such, and perhaps most importantly, to the best of our knowledge, there is no discussion of the shortcomings of standard Isolation Forest as an anomaly detection algorithm in literature. Here we present an insightful review and show the need for an improvement on the algorithm based on heat maps for anomaly scores, henceforth simply referred to as score maps. We discuss the sources of the problem in great detail, and what consequences these shortcomings bear. The importance of improving the algorithm comes to shine when we need to identify anomalies in regions where it is not immediately obvious whether a point is anomalous or not. Robust score maps are needed in assigning probabilities for given data points in terms of being anomalies.

This paper is divided into five sections. Section \ref{sec:Motivation} provides some examples that motivate this study. In section \ref{sec:EIF} we lay out the extension to the algorithm and present its application to higher dimensional datasets. In section \ref{sec:Results} we compare the results obtained by the Extended Isolation Forest to those of standard Isolation Forest, using score maps, and variance plots. Additionally, we compare rates of convergence between the two methods. We also present a comparison of AUROC and AUPRC for the two algorithms. Lastly, we close with some concluding remarks in section \ref{sec:Conclusion}. Discussion is provided throughout the various sections in the paper.

\section{Motivation} \label{sec:Motivation}
Given a dataset, we would like to be able to train our Isolation Forest so that it can assign an anomaly score for each of the data points, rank them, and help draw conclusions about the distribution of anomalies as well as the nominal points. A closer look at scores produced by Isolation Forest however, reveals that these anomaly scores are inconsistent and need some work.

\begin{figure}[H]
	\centering
	\begin{subfigure}[tc]{0.24\textwidth}
		\centering
		\includegraphics[width=0.9\linewidth]{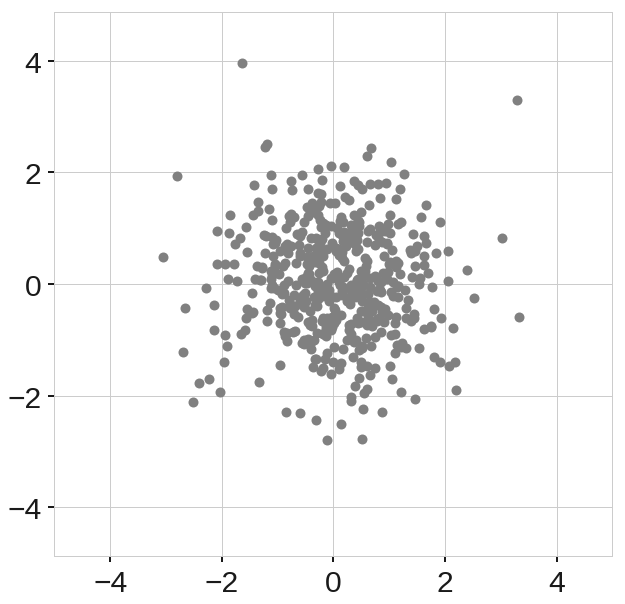}
		\caption{Normally Distributed Data}
		\label{fig:Blob_Training_Data_Background}
	\end{subfigure}
	\begin{subfigure}[tc]{0.24\textwidth}
		\centering
		\includegraphics[width=0.9\linewidth]{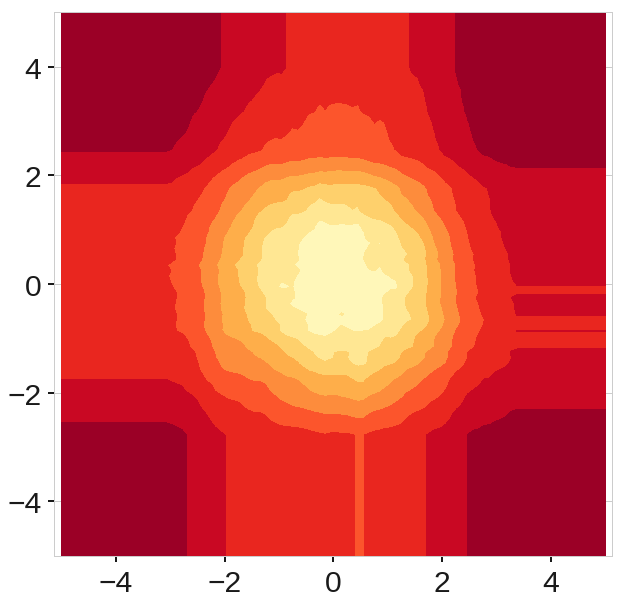}
		\caption{Anomaly Score Map}
		\label{fig:Blob_ScoreMap_Generic_BackGround}
	\end{subfigure}
	\caption{Data and anomaly score map produced by Isolation Forest for two dimensional normally distributed points with zero mean and unity covariance matrix. Darker areas indicate higher anomaly scores.}
	\label{fig:Blob_ScoreMap_Background}
\end{figure}

The best way to see the problem is to examine very simple datasets where we have an intuition of how the scores should be distributed and what constitutes an anomaly. Consider the dataset shown in Figure \ref{fig:Blob_Training_Data_Background}. This is a two dimensional random dataset sampled from a 2-D normal distribution with  zero mean vector and covariance  given by the identity matrix. It is immediately obvious that a data point should be considered nominal if it falls somewhere close to $(0,0)$. However, as the data points move away from the origin, their anomaly scores should increase. Therefore we expect to see an anomaly score map with an almost circular and symmetric pattern with increasing values as we move radially outward.

Figure \ref{fig:Blob_ScoreMap_Generic_BackGround} shows the anomaly score map obtained using the standard Isolation Forest. We sample points uniformly within the range of the plot and assign scores to those points based on the trees created for the forest.
We can clearly see that the points in the center get the lowest anomaly score, and as we move radially outward, the score values increase. However, we also observe rectangular regions of lower anomaly score in the $x$ and $y$ directions, compared to other points that fall roughly at the same radial distance from the center. Based on our understanding of how the data is distributed, the score map should maintain an approximately circular shape for all radii, i.e. similar score values for fixed distances from the origin. As we will see in later sections, the difference in the anomaly score in the $x$ and $y$ directions are indeed an artifacts introduced by the algorithm. It is critical to fix this issue for a precise and robust anomaly detection algorithm. One example of potential problems caused by this artifact is that depending on the threshold of score to label a data point an anomaly, two data points of similar importance can get categorized differently, which reduces the reliability of the algorithm. Another example might be the case where the user wants to obtain probability density functions for distribution of data points based on their anomaly scores. In this case, the results shown in Figure \ref{fig:Blob_ScoreMap_Background} can certainly not be relied upon for accurate representation of anomaly score distributions.

\begin{figure}[H]
	\centering
	\begin{subfigure}[tc]{0.24\textwidth}
		\centering
		\includegraphics[width=0.9\linewidth]{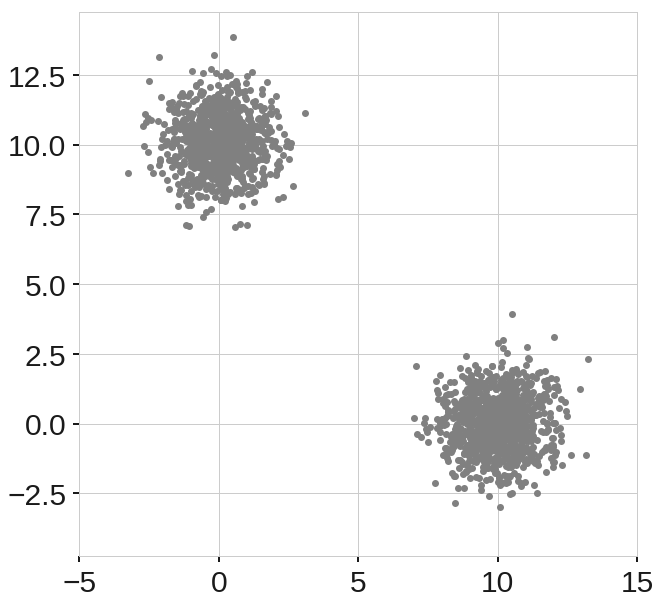}
		\caption{Two normally distributed clusters}
		\label{fig:MultipleBlobs_Training_Data_BackGround}
	\end{subfigure}
	\begin{subfigure}[tc]{0.24\textwidth}
		\centering
		\includegraphics[width=0.9\linewidth]{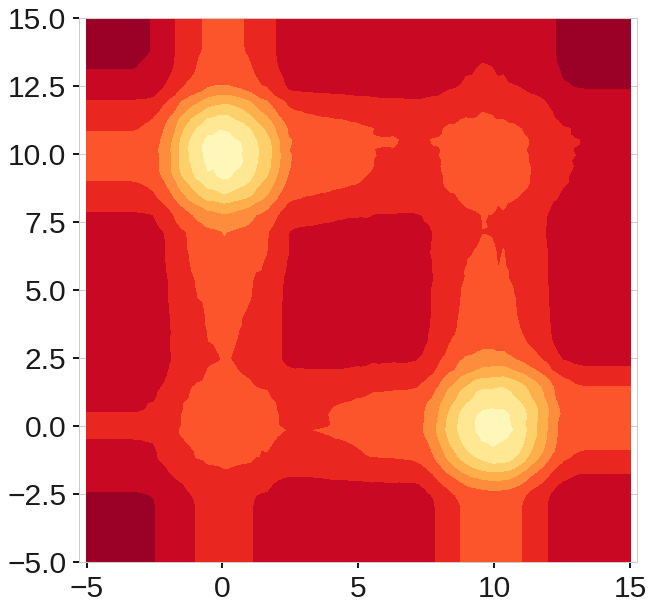}
		\caption{Anomaly Score Map}
		\label{fig:MultipleBlobs_ScoreMap_Generic_BackGround}
	\end{subfigure}
	\caption{Data points and anomaly score maps of two clusters of normally distributed points.}
	\label{fig:MultipleBlob_ScoreMap_Background}
\end{figure}

A second example is shown in Figure \ref{fig:MultipleBlob_ScoreMap_Background}. Here we have two separate clusters of normally distributed data concentrated around $(0,10)$ and $(10,0)$. We expect very low anomaly scores at the centers of the blobs, and higher scores elsewhere.

The anomaly score map clearly shows the two cluster locations and the scores distributed around them, where we can see a concentration of lower scores near the center of the clusters and higher values further away. Not surprisingly we still observe the same artifacts as in Figure \ref{fig:Blob_ScoreMap_Background} shown as rectangular bands aligned with the cluster centers. Moreover, we also see that this artifact is amplified at the intersection of the bands, close to  $(0,0)$ and $(10,10)$ where we observe ``ghost'' clusters. This introduces a real problem. In this case, if we observed an anomalous data point that happened to lie near the origin, the algorithm w5ould most likely categorize it as a nominal point. Given the nature of these artifacts it is clear that these results cannot be fully trusted for more complex data point distributions where this effect is enhanced. Not only does this increase the chances of false positives, it also wrongly indicates a non-existent structure in the data.

\begin{figure}[H]
	\centering
	\begin{subfigure}[tc]{0.24\textwidth}
		\centering
		\includegraphics[width=0.9\linewidth]{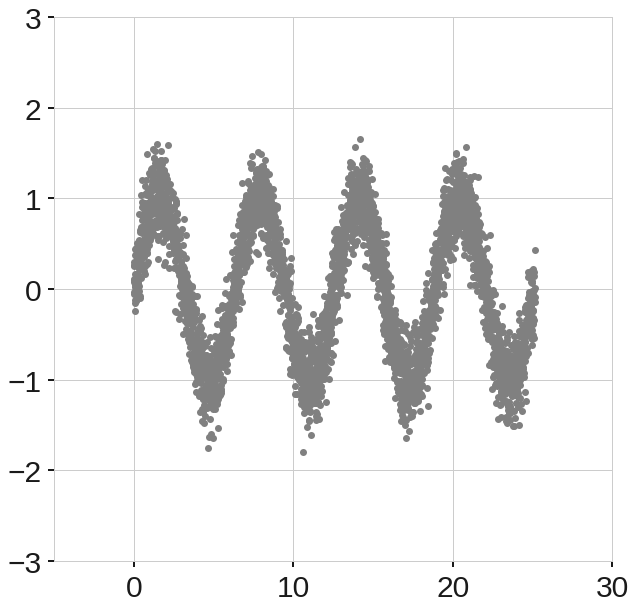}
		\caption{Sinusoidal data points with Gaussian noise.}
		\label{fig:Sinusoid_Training_Data_BackGround}
	\end{subfigure}
	\begin{subfigure}[tc]{0.24\textwidth}
		\centering
		\includegraphics[width=0.9\linewidth]{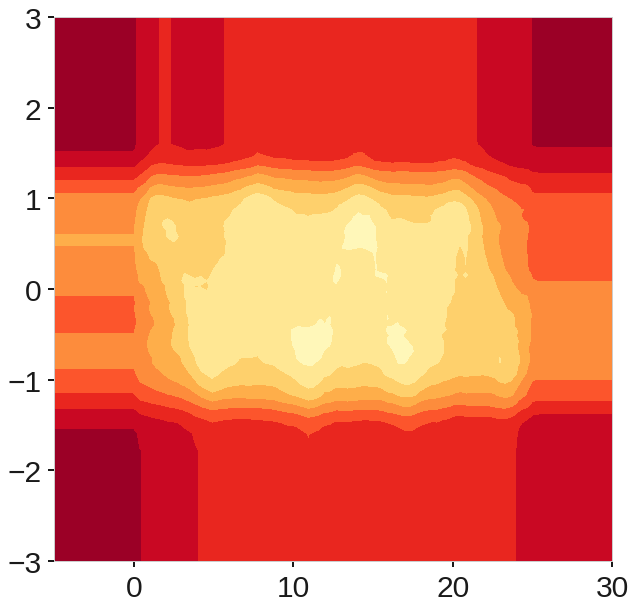}
		\caption{Anomaly Score Map}
		\label{fig:Sinusoid_ScoreMap_Generic_BackGround}
	\end{subfigure}
	\caption{Data with a sinusoidal structure and the anomaly score map.}
	\label{fig:Sinusoid_ScoreMap_Background}
\end{figure}

As a third example we will look at a dataset with more structure as shown in Figure \ref{fig:Sinusoid_ScoreMap_Background}.

In this case the data has an inherent structure, the sinusoidal shape with a Gaussian noise added on top. It is very important for the algorithm to detect this structure representing a more complex case. However, looking at the anomaly score map generated by the standard Isolation Forest, we can see that the algorithm performs very poorly. Essentially this data is treated as one large rectangular blob with horizontal and vertical bands emanating parallel to the coordinate axes as seen before. An anomalous data point that is in between two ``hills'' will get a very low score and be categorized as a nominal point.

In this work, we will discuss why these undesirable features arise, and how we can fix them using an extension of the Isolation Forest algorithm.

\section{Generalization of Isolation Forest} \label{sec:EIF}

\subsection{Overview}
\begin{figure}[H]
	\centering
	\begin{subfigure}[tc]{0.24\textwidth}
		\centering
		\includegraphics[width=0.9\linewidth]{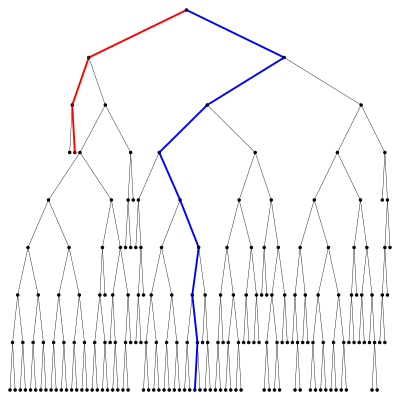}
		\caption{Representation of a single tree in a forest.}
		\label{fig:Tree}
	\end{subfigure}
	\begin{subfigure}[tc]{0.24\textwidth}
		\centering
		\includegraphics[width=0.9\linewidth]{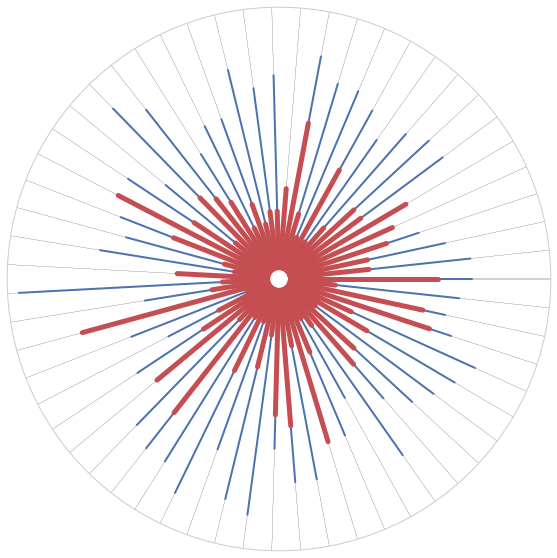}
		\caption{Representation of a full forest where each radial line corresponds to a tree.}
		\label{fig:RadialPlot_Forest}
	\end{subfigure}
	\caption{Schematic representation of a single tree (a) and a forest (b) where each tree is a radial line from the center to the outer circle. Red represents an anomaly while blue represents a nominal point.}
	\label{fig:IsoForest}
\end{figure}

The general algorithm for Isolation Forest \cite{liu2008isolation, Liu:2012:IAD:2133360.2133363} starts with the training of the data, which in this case is construction of the trees. Given a dataset of dimension $N$, the algorithm chooses a random sub-sample of data to construct a binary tree. The branching process of the tree occurs by selecting a random dimension $x_i$ with $i\in \{1,2,...,N\}$ of the data (a single variable). It then selects a random value $v$ within the minimum and maximum values in that dimension. If a given data point possesses a value smaller than $v$ for dimension $x_i$, then that point is sent to the left branch, otherwise it is sent to the right branch. In this manner the data on the current node of the tree is split in two. This process of branching is performed recursively over the dataset until a single point is isolated, or a predetermined depth limit is reached. The process begins again with a new random sub-sample to build another randomized tree. After building a large ensemble of trees, i.e. a forest, the training is complete. During the scoring step, a new candidate data point (or one chosen from the data used to create the trees) is run through all the trees, and an ensemble anomaly score is assigned based on the depth the point reaches in each tree as shown in Figure \ref{fig:IsoForest}.

Figure \ref{fig:Tree} shows a single tree after training. The red line depicts the trajectory of a single anomalous point traveling down the tree, while the blue line shows that of a nominal point. The anomalous point is isolated very quickly, but the nominal point travels all the way to the maximum depth in this case. In Figure \ref{fig:RadialPlot_Forest} we can see a full forest for an ensemble of 60 trees. Each radial line represents one tree, while the outer circle represents the maximum depth limit. The red lines are trajectories that the single anomalous point has taken down each tree, and the blue ones show trajectories of a nominal point. As can be seen, on average, the blue lines achieve a much larger radius than the red ones. It is on the basis of this idea that Isolation Forest is able to separate anomalies from nominal points.

Using the two dimensional data from Figure \ref{fig:Blob_Training_Data_Background} as a reference, during the training phase, the algorithm will create a number of a tree by picking a random dimension and comparing the value of any given point to a randomly selected cutoff value for the selected dimension. The data points are then sent down the left or the right branch according to the rules of the algorithm. By creating many such trees, we can use the average depths of the branches to assign anomaly scores. So any new observed data point can traverse down each tree following such trained rules. The average depth of the branches this data point traverses will be translated to an anomaly score using equation \eqref{eq:Score}.

\begin{align}
s(x,n) = 2^{-E(h(x))/c(n)}
	\label{eq:Score}
\end{align}
where $E(h(x))$ is the mean value of the depths a single data point, $x$,  reaches in all trees. $c(n)$ is the normalizing factor defined as the average depth in an unsuccessful search in a Binary Search Tree (BST):

\begin{align}
	c(n) = 2H(n-1) - (2(n-1)/n)
	\label{eq:c(.)}
\end{align}
where $H(i)$ is the harmonic number and can be estimated by $\ln(i) + 0.5772156649$ (Euler's constant) \cite{Liu:2012:IAD:2133360.2133363} and $n$ is the number of points used in the construction of trees.

\begin{figure}[H]
	\centering
	\begin{subfigure}[tc]{0.24\textwidth}
		\centering
		\includegraphics[width=0.9\linewidth]{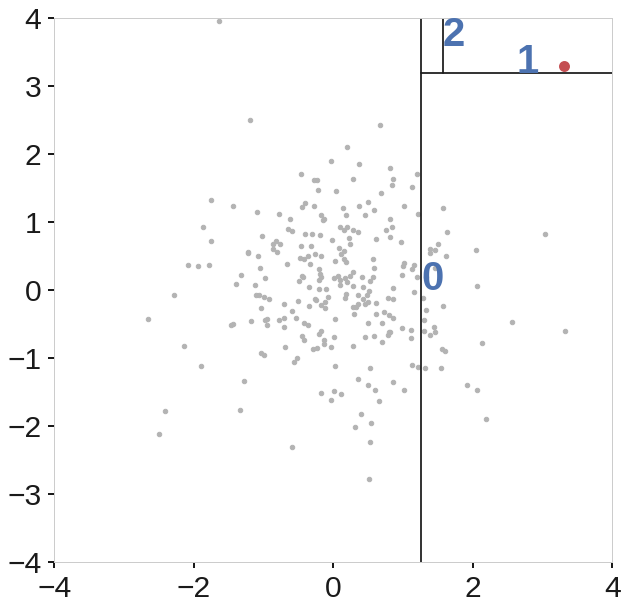}
		\caption{Anomaly point}
		\label{fig:IntersectingLines_Anomaly_Generic}
	\end{subfigure}
	\begin{subfigure}[tc]{0.24\textwidth}
		\centering
		\includegraphics[width=0.9\linewidth]{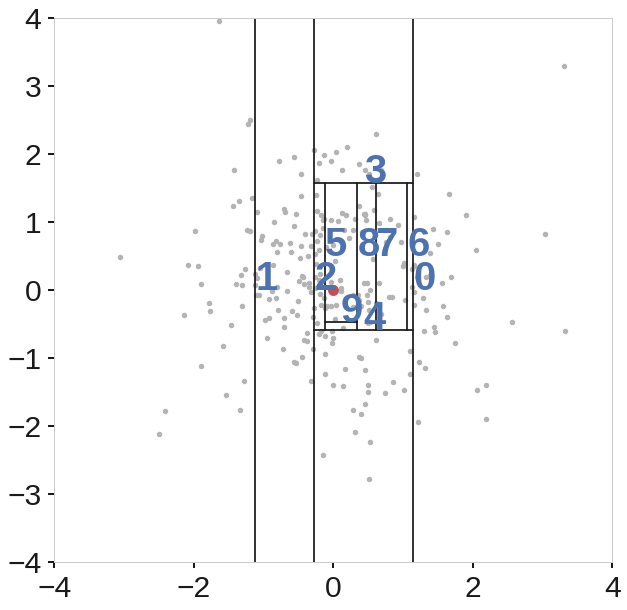}
		\caption{Nominal point}
		\label{fig:InstersectingLines_Nominal_Generic}
	\end{subfigure}
	\caption{\ref{fig:IntersectingLines_Anomaly_Generic} shows the branching process for an anomalous data point (red point). The branching takes place until the point is isolated. In this case it only took three random cuts to isolate the point. \ref{fig:InstersectingLines_Nominal_Generic} shows the same branching process for a nominal point (red point). Since the point is buried deep inside the data, it takes many cuts to isolate the point. In this case the depth limit of the tree was reached before the point was completely isolated. The numbers on the lines represent the order of branching process.}
	\label{fig:IntersectingLines_Anomaly}
\end{figure}

Trees are assigned a maximum depth limits as described in \cite{liu2008isolation}. In the case a point runs too deep in the tree, the branching process is stopped and the point is assigned the maximum depth value. Figure \ref{fig:IntersectingLines_Anomaly} shows the branching process during the training phase for an anomaly and a nominal example points using the standard Isolation Forest for the data depicted in Figure \ref{fig:Blob_ScoreMap_Background}. The numbers on the branching lines represent the order in which they were created. In the case of the nominal point, the depth limit was reached since the point is very much at the center of the data and requires many random cuts to be completely isolated. In this case the random cuts are all either vertical or horizontal. This is because of how we define the branching criteria (picking one dimension at each step). This is also the reason for the strange behavior we observed in the anomaly score maps of the previous section.

\begin{figure}[H]
	\centering
	\begin{subfigure}[tc]{0.15\textwidth}
		\centering
		\includegraphics[width=1\linewidth]{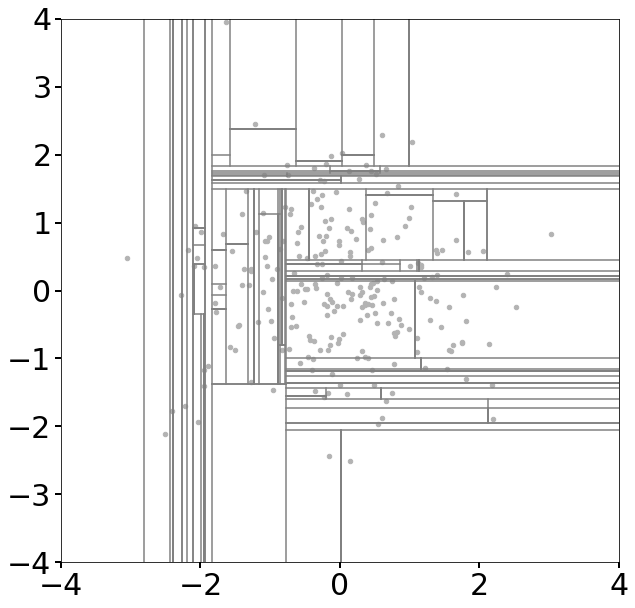}
		\caption{Single blob}
		\label{fig:SingleBlob_CutBranches_Ex0}
	\end{subfigure}
	\begin{subfigure}[tc]{0.15\textwidth}
		\centering
		\includegraphics[width=1\linewidth]{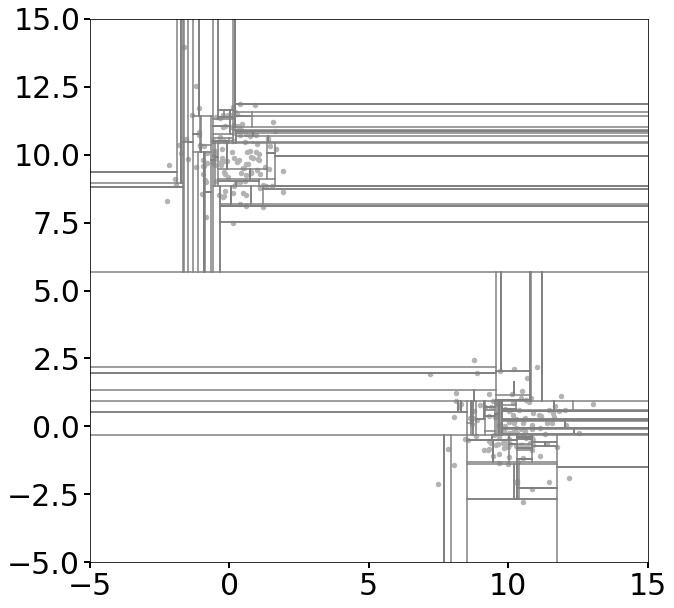}
		\caption{Multiple Blobs}
		\label{fig:MultiBlob_CutBranches_Ex0}
	\end{subfigure}
	\begin{subfigure}[tc]{0.15\textwidth}
		\centering
		\includegraphics[width=1\linewidth]{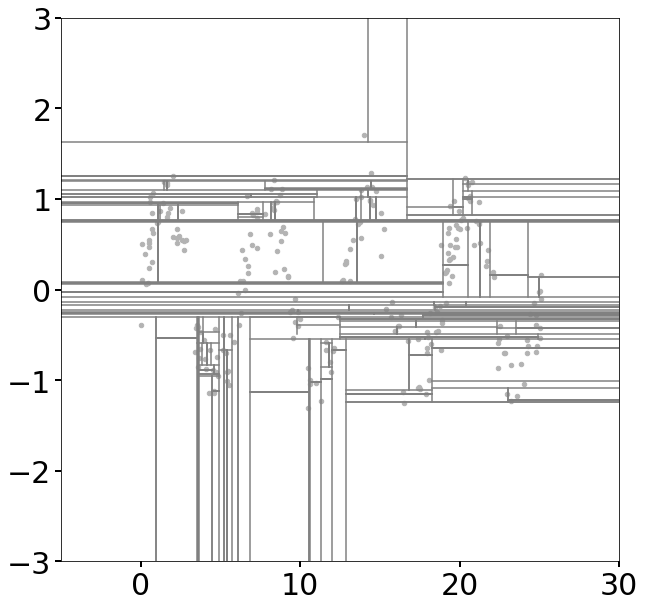}
		\caption{Sinusoidal}
		\label{fig:Sine_CutBranches_Ex0}
	\end{subfigure}
	\caption{Branch cuts generated by the standard Isolation Forest during the training phase. In each step, a random value is selected from a random feature (dimension). Then the training data points are determined to go down the left or right branches of the tree based on what side of that line they reside.}
	\label{fig:CutBranches_Ex0}
\end{figure}

To build intuition for why this is, let's consider one randomly selected, fully grown tree with all its branch rules and criteria. Figure \ref{fig:CutBranches_Ex0} shows the branch cuts generated for a tree for the three examples we have introduced in section \ref{sec:Motivation}.

\begin{figure}[H]
	\centering
	\begin{subfigure}[tc]{0.15\textwidth}
		\centering
		\includegraphics[width=1\linewidth]{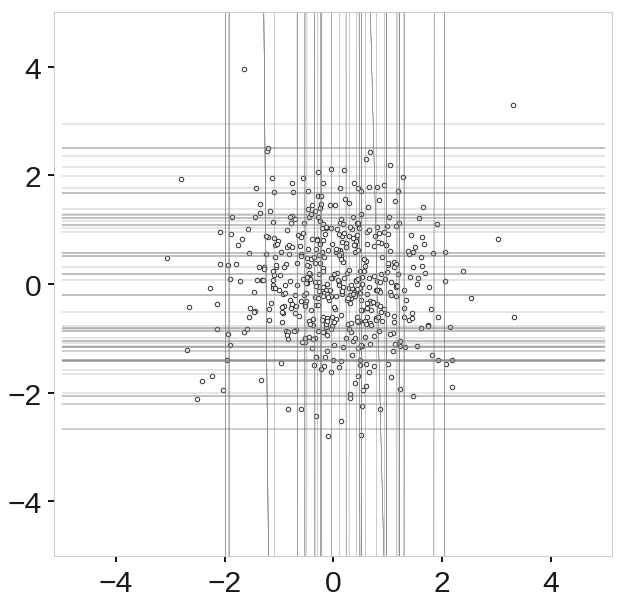}
		\caption{Single blob}
		\label{fig:AllLines_Blob_Generic}
	\end{subfigure}
	\begin{subfigure}[tc]{0.15\textwidth}
		\centering
		\includegraphics[width=1\linewidth]{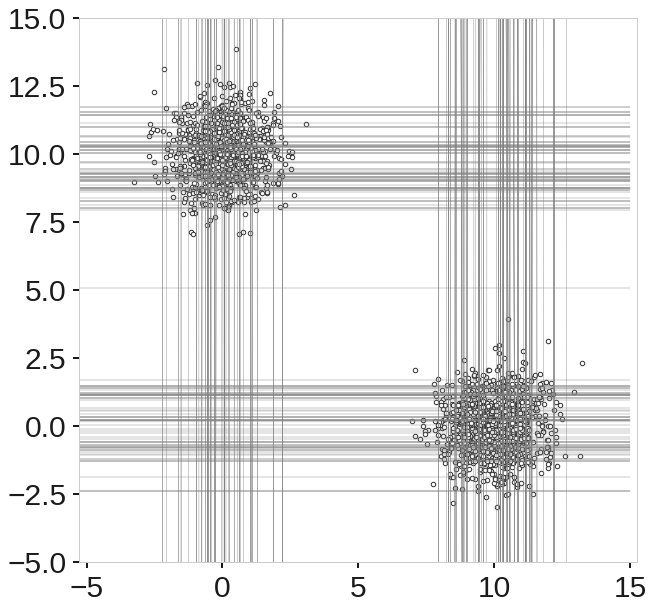}
		\caption{Multiple Blobs}
		\label{fig:AllLines_MultiplBlob_Generic}
	\end{subfigure}
	\begin{subfigure}[tc]{0.15\textwidth}
		\centering
		\includegraphics[width=1\linewidth]{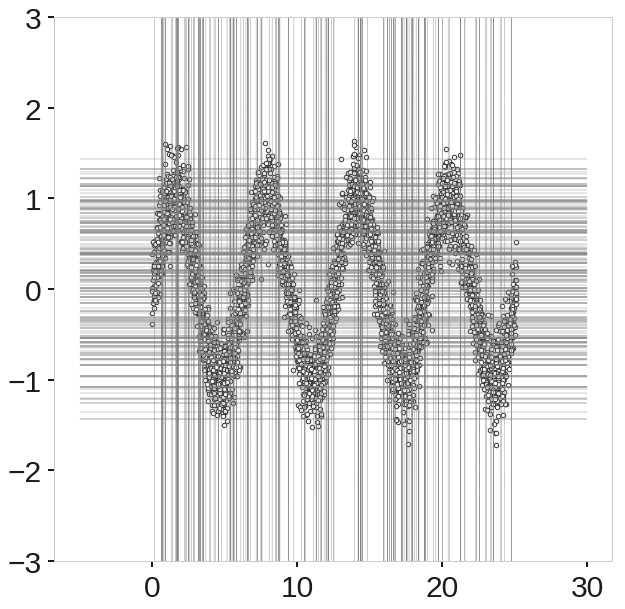}
		\caption{Sinusoidal}
		\label{fig:AllLines_Sinusoid_Generic}
	\end{subfigure}
	\caption{A typical distribution of the only possible branch cuts. The biased treatment of various regions in the domain of the data accumulate as many trees are generated, and as a result, regions of similar anomaly scores are subject to very different scoring possibilities.}
	\label{fig:AllLines_Examples_Generic}
\end{figure}

Note that in each step, we pick a random feature (dimension), $x_i$, and a random value, $v$, for this feature. Naturally, the branch cuts are simply parallel to the coordinate axes. But as we move down the branches of the tree and data is divided by these lines, the range of possible values for $v$ decreases and so the lines tend to cluster where most of the data points are concentrated. However, because of the constraint that the branch cuts are only vertical and horizontal, regions that don't necessarily contain many data points end up with many branch cuts through them. In fact Figure \ref{fig:AllLines_Examples_Generic} shows a typical distribution of the only possible branching lines (hyperplanes in general) that could possibly be produced. In Figure \ref{fig:AllLines_Blob_Generic}, a data point near $(4,0)$ will be subject to many more branching operations compared to a point that falls near $(3,3)$, despite the fact that they are both anomalies with respect to the center of the dataset. As the number of branching operations is directly related to the anomaly score, these two points might end up with very different scores.

The same phenomenon happens in the other two cases, but it is even worsened as these branch cuts intersect and form ``ghost''  cluster regions, e.g. near $(0,0)$ in \ref{fig:AllLines_MultiplBlob_Generic} as seen in previous anomaly score maps. Figure \ref{fig:AllLines_Sinusoid_Generic} shows an even more extreme case. The data repeats itself over one feature many times, and so the branch cuts in one direction are created in a much more biased fashion resulting in an inability to decipher the structure of the data as we saw in Figure \ref{fig:Sinusoid_ScoreMap_Generic_BackGround}. It is then evident that regions of similar anomalous properties receive very different branching operations throughout the training process.

We propose two different ways of addressing this limitation to provide a much more robust algorithm. In the first method, the standard Isolation Forest algorithm is allowed to run as is, but with a small modification. Each time it picks a sub-sample of the training data to create a tree, the training data has to undergo a random transformation (rotation in plane). In this case each tree has to carry the information about this transformation as it will be necessary in the scoring stage, increasing the bookkeeping. In the second method we modify and generalize the branching process by allowing the branch cuts to occur in random directions with respect the current data points on the tree node. Under this perspective, the latter method is the truly general case, while the former can be considered a special case of this general method. In fact we will show in the rest of the paper that the standard Isolation Forest is also a special case of this general method proposed here. In the following two sub-sections we explore and present each method.

\subsection{Rotated Trees}
Before we present the completely general case, we discuss rotation of the data in order to minimize the effect of the rectangular slicing. In this approach the construction of each individual tree takes place using the standard Isolation Forest, but before that happens, the sub-sample of the data that is used to construct each one of the trees, is rotated in plane, by a random angle. As such, each tree is ``twisted'' in a unique way, different than all the other trees, while keeping the structure of the data intact. Once the training is complete, and a we are computing the scores of the data points, the same transformation has to take place for each tree before the point is allowed to run down the tree branches. This approach does indeed improve the performance quite a bit as we will see in the results section.

With this method, in each case, the same bias exists as before, but only for single trees. Each tree carries with itself a different bias. When the aggregate score is computed, the total sum of the biases is averaged out resulting in a large improvement in score robustness. However, this is not the ideal solution for fixing problems presented above, despite the fact that it seems to produce better results. Some reasons as to why this method is less desirable are:

\begin{enumerate}
	\item Each tree has to be tagged with its unique rotation so that when we are scoring observed data, we can compensate for the rotation in the coordinates of the data point.
	\item Even though the ensemble results seem good, each tree still suffers from the rectangular bias introduced by the underlying algorithm. In a sense the problem is not resolved, but only averaged out.
	\item This approach can become cumbersome to apply especially with large datasets and higher dimensions.
	\item It is prone to simple errors if datasets lack symmetries.
	\item The rotation is not obvious in higher dimensions than 2-D. For each tree we can pick a random axis in the space and perform planar rotation around that axis, but there are many other choices that can be made, which might result in inconsistencies among different runs.
	\item There is extra bookkeeping and meta data store for each tree.
\end{enumerate}

A much more robust fix to the problem can be achieved by truly randomizing the branching process in addition to the randomization already in place. That is, instead of simply running the Isolation Forest for each tree using a rotated sub-sample, we select different angles for the data slices at each branching point.

\subsection{Extended Isolation Forest}

\begin{figure}[H]
	\centering
	\begin{subfigure}[tc]{0.24\textwidth}
		\centering
		\includegraphics[width=1\linewidth]{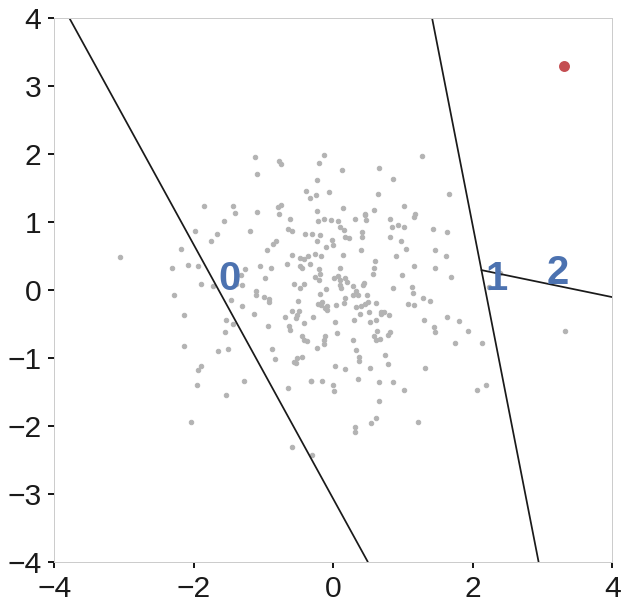}
		\caption{Anomaly}
		\label{fig:IntersectingLines_Anomaly_Extended}
	\end{subfigure}
	\begin{subfigure}[tc]{0.24\textwidth}
		\centering
		\includegraphics[width=1\linewidth]{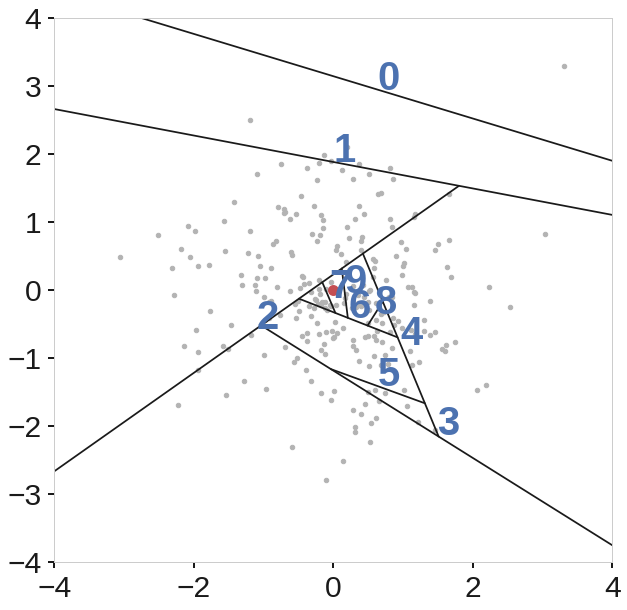}
		\caption{Nominal}
		\label{fig:IntersectingLines_Nominal_Extended}
	\end{subfigure}
	\caption{Branching process for the Extended Isolation Forest. Figure \ref{fig:IntersectingLines_Anomaly_Extended} shows the branching process for an anomalous data point. The branching takes place until the point is isolated. In this case it only took three random cuts to isolate the point. \ref{fig:IntersectingLines_Nominal_Extended} shows the same branching process for a nominal point. Since the point is near the center of the data, it takes many cuts to isolate the point. In this case the depth limit of the tree was reached before the point was isolated.}
	\label{fig:IntersectingLines_Extended}
\end{figure}

Isolation Forest relies on randomness in selection of features and values. Since anomalous points are ``few and different'', they quickly stand out with respect to these random selections. But as we have seen, the branch cuts are always either horizontal or vertical, and this introduces a bias and artifacts in the anomaly score map. There is no fundamental reason in the algorithm that requires this restriction, and so at each branching point, we can select a branch cut that has a random ``slope''. A single branching process for branch cuts with random slopes in our 2-D example is visualized in Figure \ref{fig:IntersectingLines_Extended} for both cases of anomaly and nominal data points, analogous to Figure \ref{fig:IntersectingLines_Anomaly}.

Similarly to the standard Isolation Forest algorithm, anomalies can be isolated very quickly, whereas the nominal points require many branch cuts to be isolated. In the case of the standard Isolation Forest algorithm, the selection of the branch cuts requires two pieces of information: 1) a random feature or coordinate, and 2) a random value for the feature from the range of available values in the data. For the extended case, the selection of the branch cuts still only requires two pieces of information, but they are: 1) a random slope for the branch cut, and 2) a random intercept for the branch cut which is chosen from the range of available values of the training data. Notice that this is simpler than the tree rotations in the previous section, because in that case in addition to the two pieces of information at each branch cut, each tree needs to store the information for transforming the data so that it can be used in the scoring stage.

For an $N$ dimensional dataset, selecting a random slope for the branch cut is the same as choosing a normal vector, $\vec{n}$, uniformly over the unit $N$-Sphere. This can easily be accomplished by drawing a random number for each coordinate of $\vec{n}$ from the standard normal distribution $\mathcal{N}(0,1)$ \cite{harman2010decompositional}. This results in a uniform selection of points on the $N$-sphere. For the intercept, $\vec{p}$, we simply draw from a uniform distribution over the range of values present at each branching point. Once these two pieces of information are determined, the branching criteria for the data splitting for a given point $\vec{x}$ is as follows:

\begin{align}
\left( \vec{x} - \vec{p} \right) \cdot \vec{n} \leq 0
\label{eq:BranchingTest}
\end{align}

If the condition is satisfied, the data point $\vec{x}$ is passed to the left branch, otherwise it moves down to the right branch.

\begin{figure}[H]
	\centering
	\begin{subfigure}[tc]{0.15\textwidth}
		\centering
		\includegraphics[width=1\linewidth]{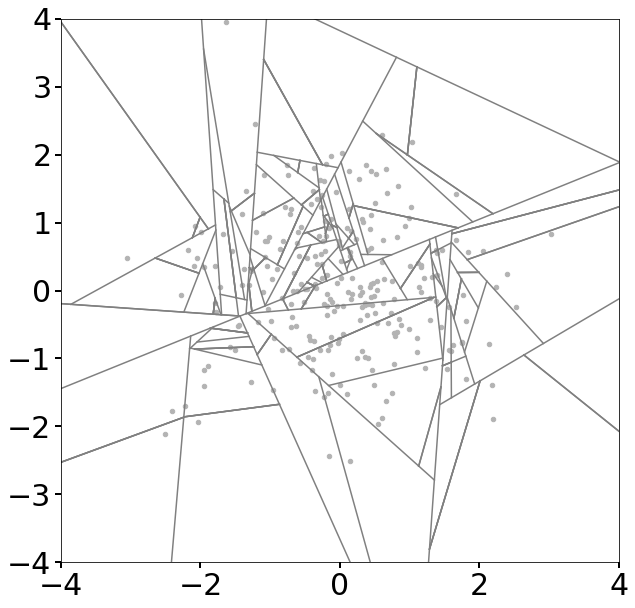}
		\caption{Single blob}
		\label{fig:SingleBlob_CutBranches_Ex1}
	\end{subfigure}
	\begin{subfigure}[tc]{0.15\textwidth}
		\centering
		\includegraphics[width=1\linewidth]{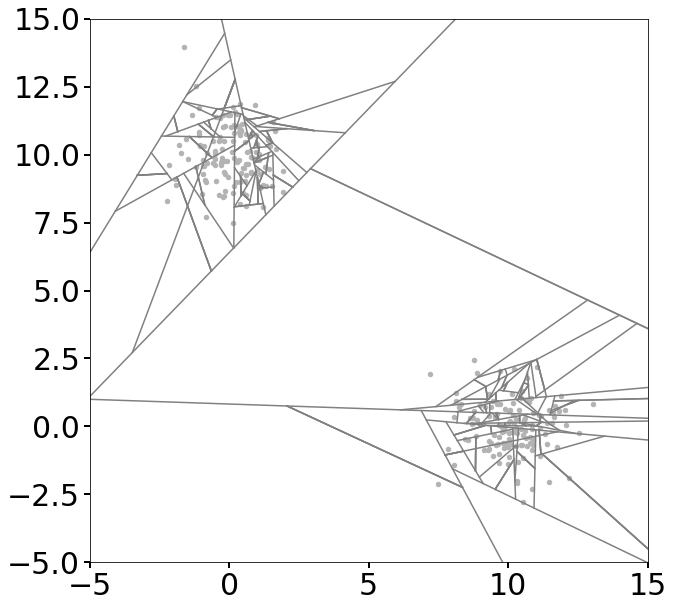}
		\caption{Multiple Blobs}
		\label{fig:MultiBlob_CutBranches_Ex1}
	\end{subfigure}
	\begin{subfigure}[tc]{0.15\textwidth}
		\centering
		\includegraphics[width=1\linewidth]{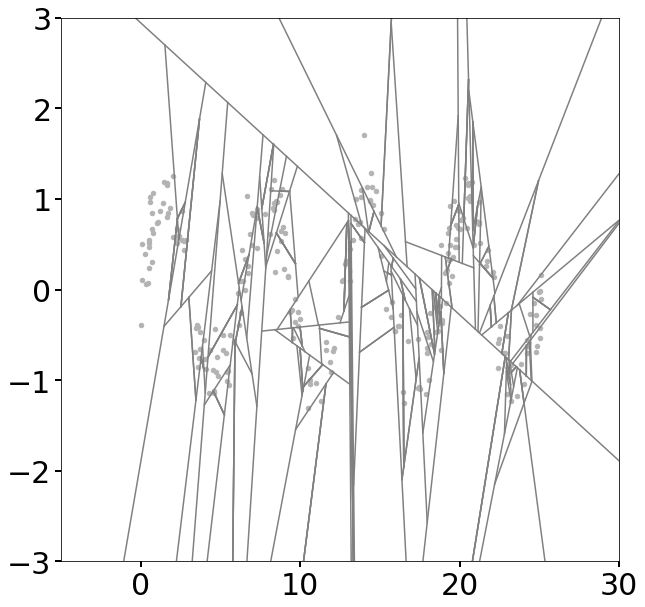}
		\caption{Sinusoidal}
		\label{fig:Sine_CutBranches_Ex1}
	\end{subfigure}
	\caption{Branch cuts generated by the Extended Isolation Forest. In each step, a normal vector, $\vec{n}$, along with a random intercept point, $\vec{p}$, is selected. A data point, $\vec{x}$, is determined to go down the left or right branches of the tree based on the criteria shown in inequality \eqref{eq:BranchingTest}.}
	\label{fig:CutBranches_Ex1}
\end{figure}

As we saw in Figure \ref{fig:IntersectingLines_Extended}, the branch cuts are drawn with random slopes until they isolate the data points, or until the depth limit is reached. Figure \ref{fig:CutBranches_Ex1} is analogous to Figure \ref{fig:CutBranches_Ex0}. The branch cuts are shown for training of a single tree. We note that despite the randomness of the process, higher density points are better represented schematically in comparison with Figure \ref{fig:CutBranches_Ex0}.

\begin{figure}[H]
	\centering
	\begin{subfigure}[tc]{0.15\textwidth}
		\centering
		\includegraphics[width=1\linewidth]{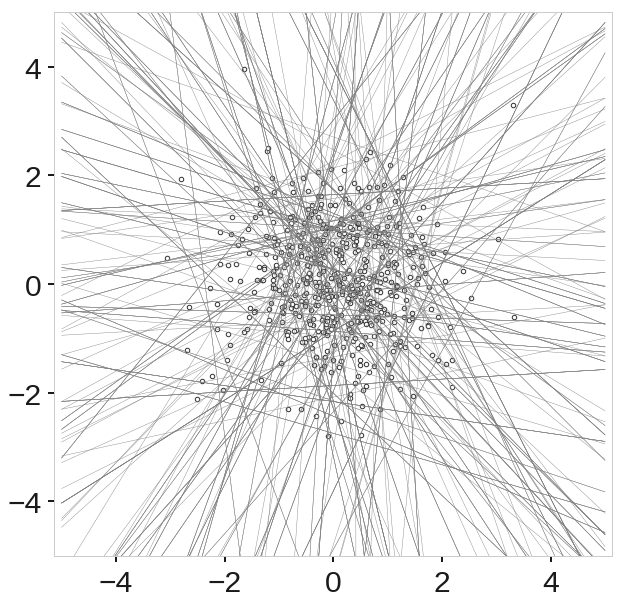}
		\caption{Single blob}
		\label{fig:AllLines_Blob_Extended}
	\end{subfigure}
	\begin{subfigure}[tc]{0.15\textwidth}
		\centering
		\includegraphics[width=1\linewidth]{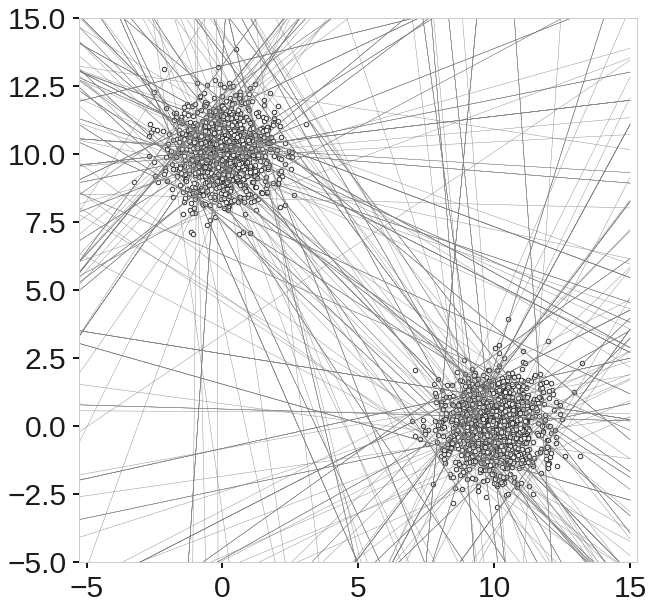}
		\caption{Multiple Blobs}
		\label{fig:AllLines_MultiplBlob_Extended}
	\end{subfigure}
	\begin{subfigure}[tc]{0.15\textwidth}
		\centering
		\includegraphics[width=1\linewidth]{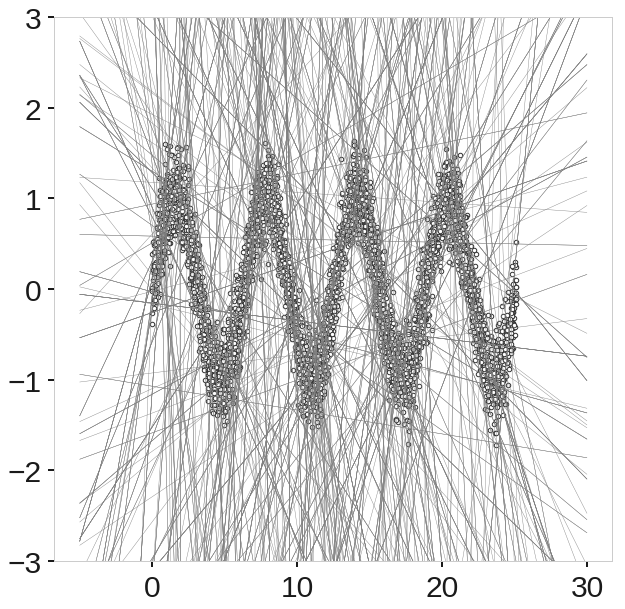}
		\caption{Sinusoid}
		\label{fig:AllLines_Sinusoid_Extended}
	\end{subfigure}
	\caption{A typical distribution of the possible branch cuts for the case of Extended Isolation Forest. The branch cuts produced can now pass through any part of the region free of biases introduced by the coordinate axes constraints.}
	\label{fig:AllLines_Examples_Extended}
\end{figure}

Figure \ref{fig:AllLines_Examples_Extended} shows typical plots of all the possible branch cuts that can be produced by the Extended Isolation Forest for each of the examples we have considered thus far, similar to the figures shown in \ref{fig:AllLines_Examples_Generic}. Unlike the previous case, we can clearly see that branch cuts are possible in all regions of the domain regardless of the coordinate axes.

The important property of concentrating where the data is clustered, persists. The intercept points $\vec{p}$ tend to accumulate where the data is, since the selection of these points gets restricted to available data at each branching point as we move deeper into the tree. This results in more possible branching options where the data is clustered and fewer possible branching where there is less concentration of data. However, there are no regions that artificially receive more attention than the rest. The results of this are score maps that are free of artifacts previously observed.

\subsection{High Dimensional Data and Extension Levels}

So far we have only been looking at two dimensional data because they are easy to visualize. However, the algorithm generalizes readily to higher dimensions. In this case, the branch cuts are no longer straight line, but $N-1$ dimensional hyperplanes. These hyperplanes can still be specified with a random normal vector $\vec{n}$ and a random intercept point $\vec{p}$ whose selection are identical to that of the two dimensional case. The same criteria for branching process specified by inequality \eqref{eq:BranchingTest} applies to the high dimensional case.

It is interesting to note that for an $N$ dimensional dataset we can consider $N$ levels of extension. In the fully extended case, we select our normal vector by drawing each component from  $\mathcal{N}(0,1)$ as seen before. This results in hyperplanes that can intersect any of the coordinates axes. For example in the case of two dimensions, it resulted in oblique lines. However, we can exclude exactly one dimension in specifying the lines so that they are parallel to one of the two coordinate axes. This is simply accomplished by setting a coordinate of the normal vector to zero, in which case we recover the standard Isolation Forest.

\begin{figure}[H]
	\centering
	\begin{subfigure}[tc]{0.15\textwidth}
		\centering
		\includegraphics[width=1\linewidth]{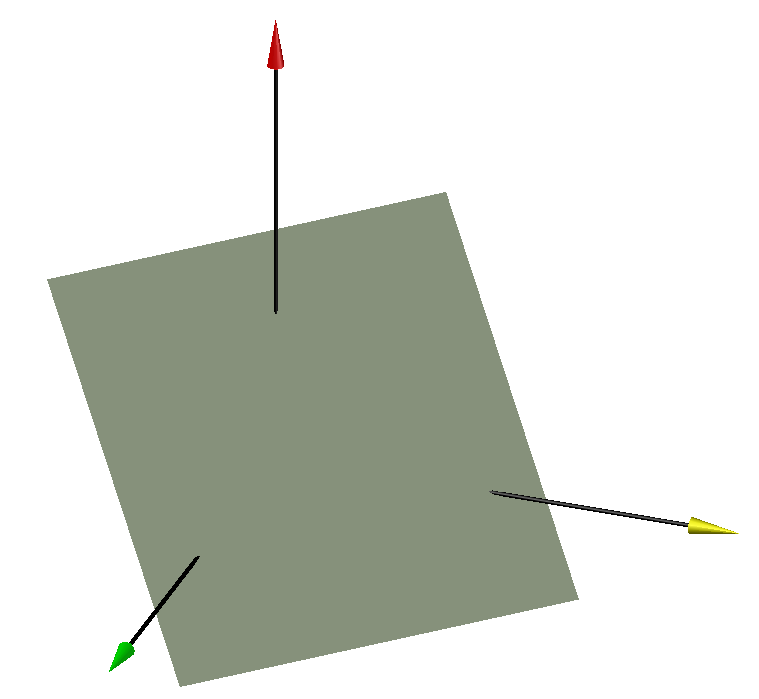}
		\caption{Ex 2}
		\label{fig:general2}
	\end{subfigure}
	\begin{subfigure}[tc]{0.15\textwidth}
		\centering
		\includegraphics[width=1\linewidth]{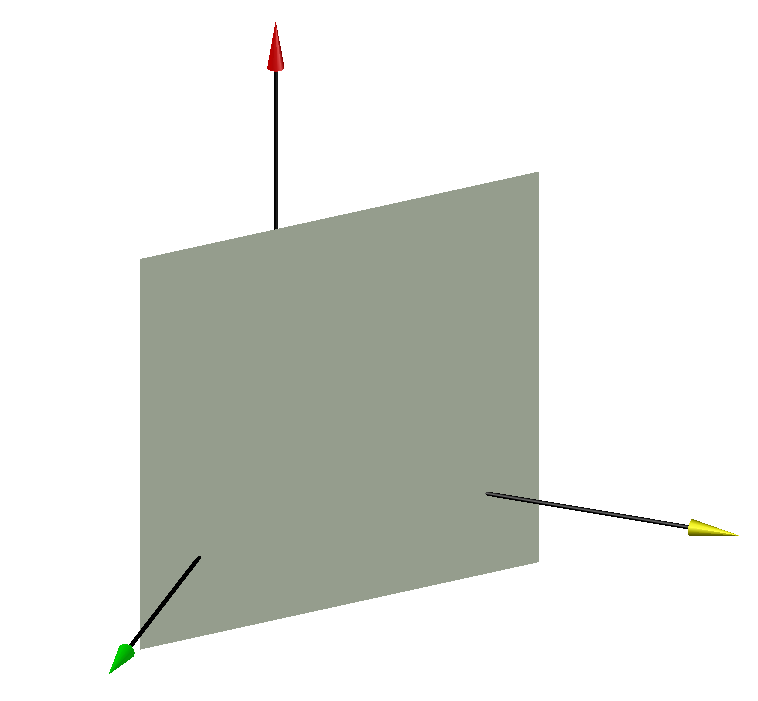}
		\caption{Ex 1}
		\label{fig:general1}
	\end{subfigure}
	\begin{subfigure}[tc]{0.15\textwidth}
		\centering
		\includegraphics[width=1\linewidth]{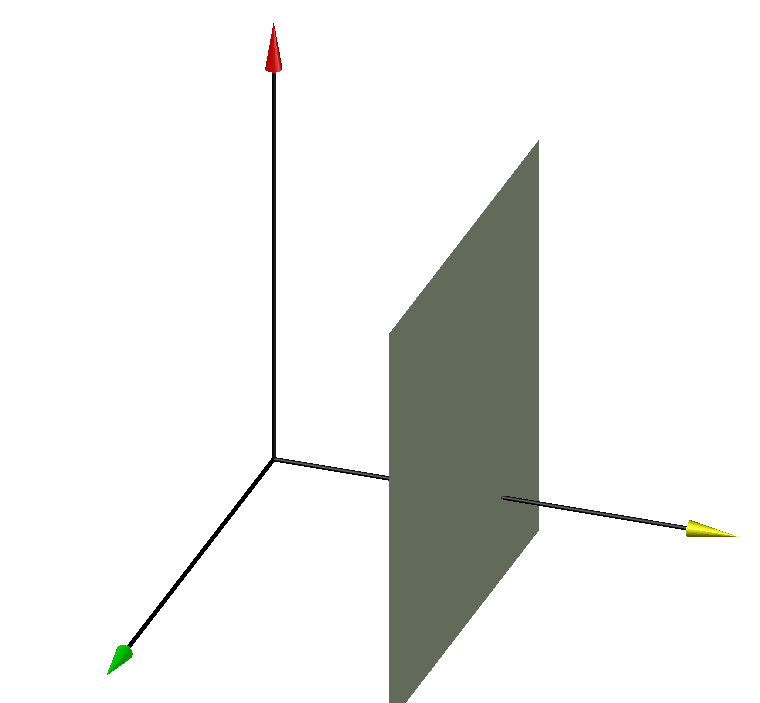}
		\caption{Ex 0}
		\label{fig:general0}
	\end{subfigure}
	\caption{Sample branch hyperplane for three dimensional data for each level of extension. The case of extension 0 \ref{fig:general0} is identical to the standard Isolation Forest.}
	\label{fig:general}
\end{figure}

In the case of three dimensional data for example, the fully extended case is when the normal vectors are selected so that the hyperplanes (in this case 2-D planes) are allowed to intersect with all three axes. We call that the fully extended case or \nth{2} Extension (Ex 2). If we reduce the extension level by one, the 2-D planes are always parallel to one of the three coordinates, and we call that \nth{1} Extension (Ex 1). Reducing the extension yet again, we arrive at \nth{0} extension, which is the case where the random slices are always parallel to two of the axes. This case is identically equivalent to the standard Isolation Forest algorithm. At each branching step we essentially have one active coordinate or feature over which a random value is selected to slice the training data. See Figure \ref{fig:general}

So for any given $N$ dimensional dataset, the lowest level of extension of the Extended Isolation Forest is coincident with the standard Isolation Forest. As the extension level of the algorithm is increased, the bias of the algorithm is reduced, see Figure \ref{fig:STDvsRadius_4D} for an example. The idea of having multiple levels of extension can be useful in cases where the dynamic range of the data in various dimensions are very different. In such cases, reducing the extension level can help in more appropriate selection of split hyperplanes and reducing the computational overhead. As an extreme case, if we had three dimensional data, but the range in two of the dimensions were much smaller compared to the third (essentially data distributed along a line), the standard Isolation Forest method would probably yield the most optimal result.

\subsection{The Algorithm}

The are few changes to the original algorithm, published in \cite{liu2008isolation} which are presented here. Algorithm \ref{alg:iForest} remains the same which we show for completeness.

\begin{center}
	\begin{minipage}{.9\linewidth}
		\begin{algorithm}[H]
			\caption{$iForest(X, t, \psi$)}
			\begin{algorithmic}[1]
				\Require $X$ - input data, $t$ - number of trees, $\psi$ - sub-sampling size
				\Ensure a set of $t$ $iTrees$

				\State \textbf{Initialize} $Forests$
				\State set height limit $l=ceiling(\log_2 \psi)$
				\For{$i=1$ to $t$}
				\State $X' \gets sample(X, \psi)$
				\State $Forest \gets Forest \cup iTree(X',0,l)$
				\EndFor
			\end{algorithmic}
			\label{alg:iForest}
		\end{algorithm}
	\end{minipage}
\end{center}

As mentioned before, the Extended Isolation Forest algorithm only requires determining two random pieces of information at the branching nodes: the normal vector, and the intercept point. In algorithm \ref{alg:iTree}, we modified the two lines that pick a random feature and a random value for that feature with lines 4 and 5. In addition to that, we also change the test condition to reflect inequality \eqref{eq:BranchingTest}.  In addition, we have added line 6 which has to do with allowing the extension level to change. With these changes, the algorithm can be used as either the standard Isolation Forest, or as the Extended Isolation Forest with any desired extension level. This extension customization can be useful in cases, for example, of a complex multidimensional data with low cardinality in one dimension where that feature can be ignored to reduce the introduction of possible selection bias.

\begin{center}
	\begin{minipage}{.9\linewidth}
		\begin{algorithm}[H]
			\caption{$iTree(X, e, l$)}
			\begin{algorithmic}[1]
				\Require $X$ - input data, $e$ - current tree height, $l$ - height limit
				\Ensure an iTree
				\If{$e \geq l$ or $|X| \leq 1$}
				\State \Return $exNode\{Size \gets |X|\}$
				\Else
				\State randomly select a normal vector $\vec{n} \in {\rm I\!R}^{|X|}$ by drawing each coordinate of $\vec{n}$ from a standard Gaussian distribution.
				\State randomly select an intercept point $\vec{p} \in  {\rm I\!R}^{|X|}$ in the range of $X$
				\State set coordinates of $\vec{n}$ to zero according to extension level
				\State $X_l \gets filter(X,(X-\vec{p})\cdot \vec{n} \leq 0)$
				\State $X_r \gets filter(X,(X-\vec{p})\cdot \vec{n} > 0)$
				\State \Return inNode$\{
				\begin{aligned}[t]
				& Left \gets iTree(X_l,e+1, l), \\
				& Right \gets iTree(X_r,e+1,l), \\
				& Normal \gets \vec{n}, \\
				& Intercept \gets \vec{p}\}
				\end{aligned}$
				\EndIf
			\end{algorithmic}
			\label{alg:iTree}
		\end{algorithm}
	\end{minipage}
\end{center}

In algorithm \ref{alg:PathLength}, we make the changes accordingly. We now have to receive the normal and intercept point from each tree, and use the appropriate test condition to set off the recursion for figuring out path length (depth of each branch).

\begin{center}
	\begin{minipage}{.9\linewidth}
		\begin{algorithm}[H]
			\caption{$PathLength(\vec{x},T,e)$}
			\begin{algorithmic}[1]
				\Require $\vec{x}$ - an instance, $T$ - an iTree, $e$ - current path length; to be initialized to zero when first called
				\Ensure path length of $\vec{x}$
				\If{$T$ is an external node}
				\State \Return $e + c(T.size)\{c(.) \text{ is defined in Equation \eqref{eq:c(.)}}\}$
				\EndIf
				\State $\vec{n} \gets T.Normal$
				\State $\vec{p} \gets T.Intercept$
				\If {$(\vec{x}-\vec{p})\cdot \vec{n} \leq 0$}
				\State \Return $PathLength(\vec{x},T.left, e+1)$
				\ElsIf {$(\vec{x}-\vec{p})\cdot \vec{n} > 0$}
				\State \Return $PathLength(\vec{x},T.rigth, e+1)$
				\EndIf
			\end{algorithmic}
			\label{alg:PathLength}
		\end{algorithm}
	\end{minipage}
\end{center}

As we can see the algorithm can easily be modified to take into account all the necessary changes for this extension. We have made publicly available a Python implementation of these algorithms \footnote{https://github.com/sahandha/eif} accompanied by example Jupyter notebooks to reproduce the figures in the text.

\section{Results and Discussion} \label{sec:Results}

Here we present the results of the Extended Isolation Forest compared to the standard Isolation Forest. We compare the anomaly score maps, variance plots and convergence of the anomaly scores. We also report on the AUROC and AUPRC values in order to quantify the comparison between the two methods.

\subsection{Score Maps}

We first look at the score maps of the two dimensional examples presented in section \ref{sec:Intro}.
The score maps of these examples were the initial motivation for improving and extending the algorithm. Note that these plots are only possible for two dimensional data, even though it is possible to visualize these maps and artifacts in 3-D. We will use different metrics in order to analyze higher dimensional data.

\begin{figure}[H]
	\centering
	\begin{subfigure}[tc]{0.15\textwidth}
		\centering
		\includegraphics[width=1\linewidth]{./Images/Blob_ScoreMap_Generic.png}
		\caption{Standard IF}
		\label{fig:Blob_ScoreMap_Generic_Result}
	\end{subfigure}
	\begin{subfigure}[tc]{0.15\textwidth}
		\centering
		\includegraphics[width=1\linewidth]{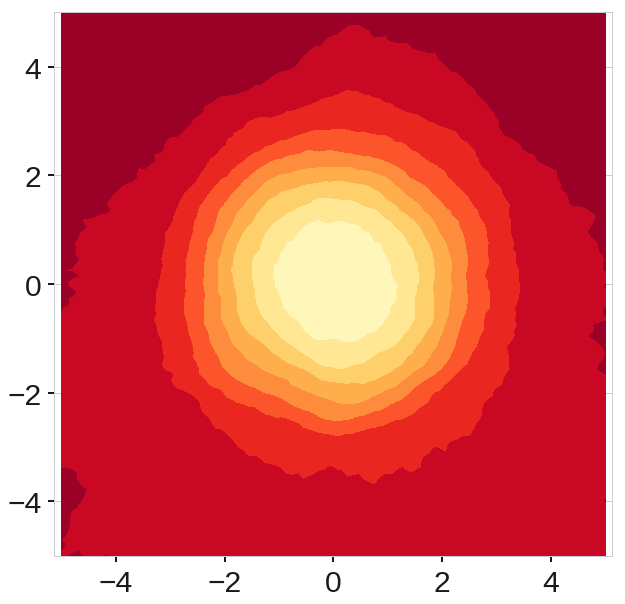}
		\caption{Rotated IF}
		\label{fig:Blob_ScoreMap_Rotated_Result}
	\end{subfigure}
	\begin{subfigure}[tc]{0.15\textwidth}
		\centering
		\includegraphics[width=1\linewidth]{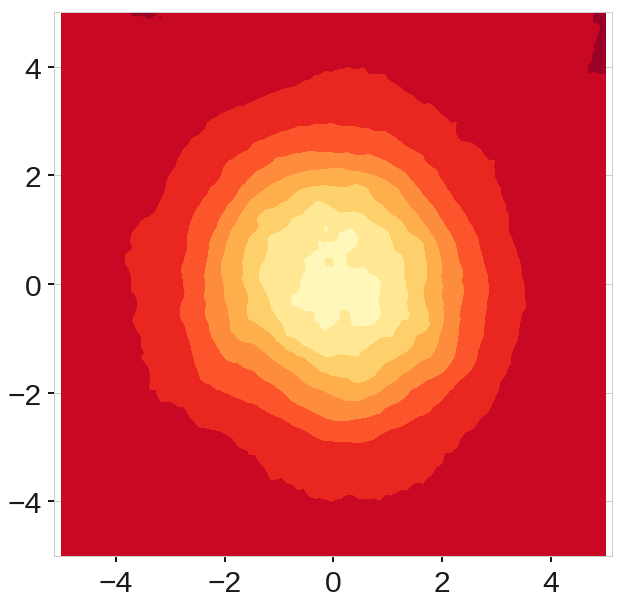}
		\caption{Extended IF}
		\label{fig:Blob_ScoreMap_Extended_Result}
	\end{subfigure}
	\caption{Comparison of the Standard Isolation Forest, Rotated Isolation Forest, and Extended Isolation Forest for the case of the single blob.}
	\label{fig:Blob_ScoreMap_Result}
\end{figure}

First consider the case of a single blob normally distributed in 2-D space. Figure \ref{fig:Blob_ScoreMap_Result} compares the anomaly score maps for all the mentioned methods.

By comparing Figure \ref{fig:Blob_ScoreMap_Generic_Result} to \ref{fig:Blob_ScoreMap_Rotated_Result} we can already see a considerable difference in the anomaly score map. The rotation of the sub-sampled data for each tree averages out the artifacts and produces a more symmetric map as  expected in this example.
Taking it a step further, we look at the score map in \ref{fig:Blob_ScoreMap_Extended_Result}, which was obtained by the Extended Isolation Forest. We can see an anomaly score map as we might  expect for this example. The low score artificial bands in the $x$ and $y$ direction are no longer present and the score increases roughly monotonically in all directions as we move radially outward from the center in a quasi-symmetric way.

\begin{figure}[H]
	\centering
	\begin{subfigure}[tc]{0.15\textwidth}
		\centering
		\includegraphics[width=1\linewidth]{./Images/MultipleBlobs_ScoreMap_Generic.png}
		\caption{Standard IF}
		\label{fig:MultipleBlobs_ScoreMap_Generic_Result}
	\end{subfigure}
	\begin{subfigure}[tc]{0.15\textwidth}
		\centering
		\includegraphics[width=1\linewidth]{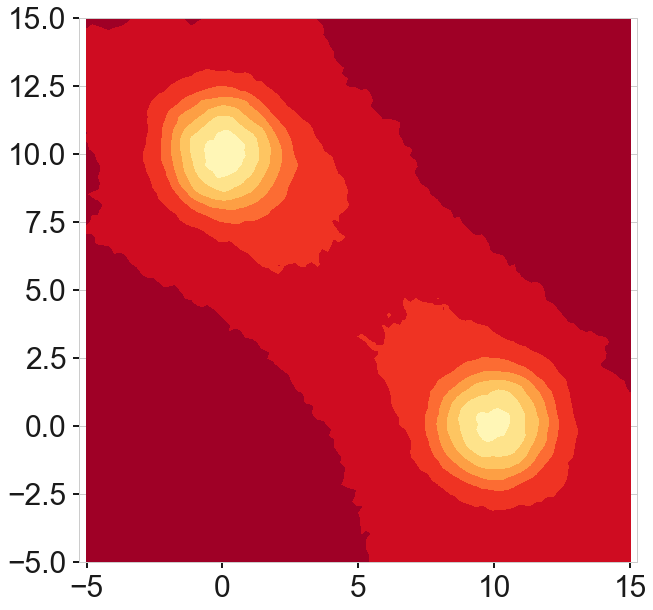}
		\caption{Rotated IF}
		\label{fig:MultipleBlobs_ScoreMap_Rotated_Result}
	\end{subfigure}
	\begin{subfigure}[tc]{0.15\textwidth}
		\centering
		\includegraphics[width=1\linewidth]{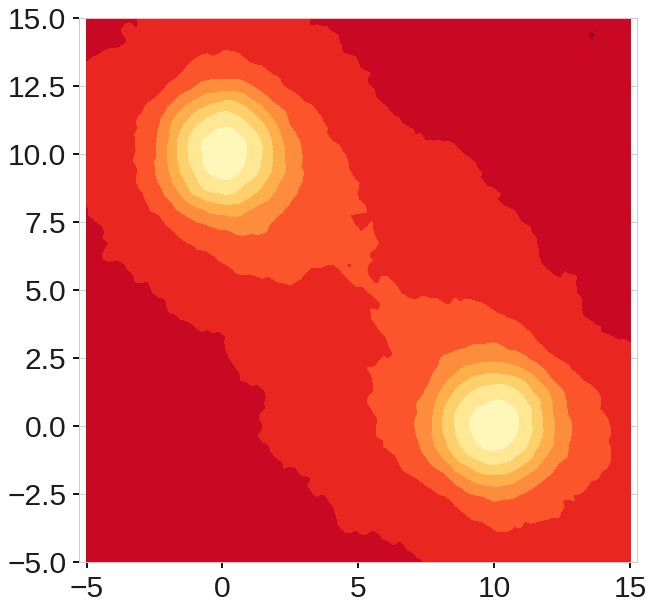}
		\caption{Extended IF}
		\label{fig:MultipleBlobs_ScoreMap_Extended_Result}
	\end{subfigure}
	\caption{Comparison of the standard Isolation Forest with rotated Isolation Forest, and Extended Isolation Forest for the case of two blobs.}
	\label{fig:MultipleBlobs_ScoreMap_Result}
\end{figure}

Now we turn to the case of multiple blobs presented in Figure \ref{fig:MultipleBlobs_ScoreMap_Result}. As we saw for the single blob case and as shown in Figure \ref{fig:MultipleBlobs_ScoreMap_Generic_Result}, the creation of branch cuts that were only parallel to the coordinate axes was causing two ``ghost'' regions in the score map, in addition to the bands observed in the single blob case.

As shown in Figures \ref{fig:MultipleBlobs_ScoreMap_Rotated_Result} and \ref{fig:MultipleBlobs_ScoreMap_Extended_Result}, in both cases, namely, the Rotated and Extended Isolation Forest, these ``ghost'' regions are completely gone. The two blobs are clearly scored with low anomaly scores, as expected. The interesting feature to notice here is the higher anomaly score region directly in between the two blobs. The Extended Isolation Forest algorithm is able to capture this detail quite well. This additional feature is important as this region can be considered as close to nominal given the proximity to the blobs, but with higher score since it is far from the concentrated distribution. The Standard Isolation Forest fails in capturing this extra structure from the data.

As for the sinusoidal data, we observed before that the standard Isolation Forest failed to detect the structure of the data and treated it as one rectangular blob with very wide rectangular bands.

\begin{figure}[H]
	\centering
	\begin{subfigure}[tc]{0.15\textwidth}
		\centering
		\includegraphics[width=1\linewidth]{./Images/Sinusoid_ScoreMap_Generic.png}
		\caption{Generic IF}
		\label{fig:Sinusoid_ScoreMap_Generic_Result}
	\end{subfigure}
	\begin{subfigure}[tc]{0.15\textwidth}
		\centering
		\includegraphics[width=1\linewidth]{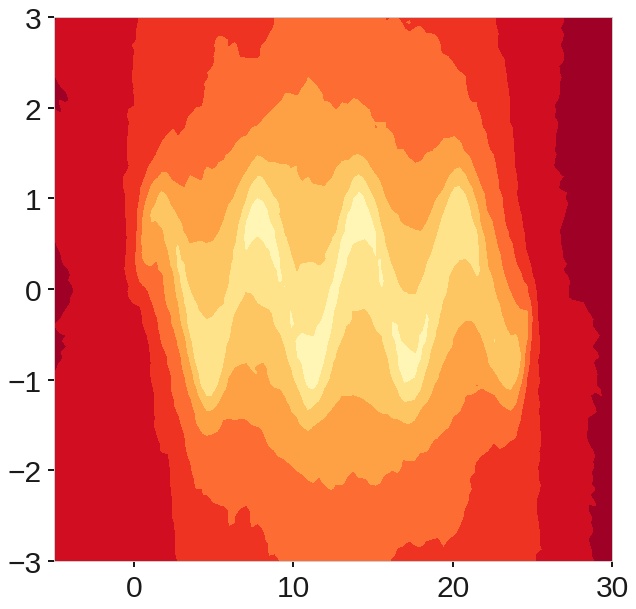}
		\caption{Rotated IF}
		\label{fig:Sinusoid_ScoreMap_Rotated_Result}
	\end{subfigure}
	\begin{subfigure}[tc]{0.15\textwidth}
		\centering
		\includegraphics[width=1\linewidth]{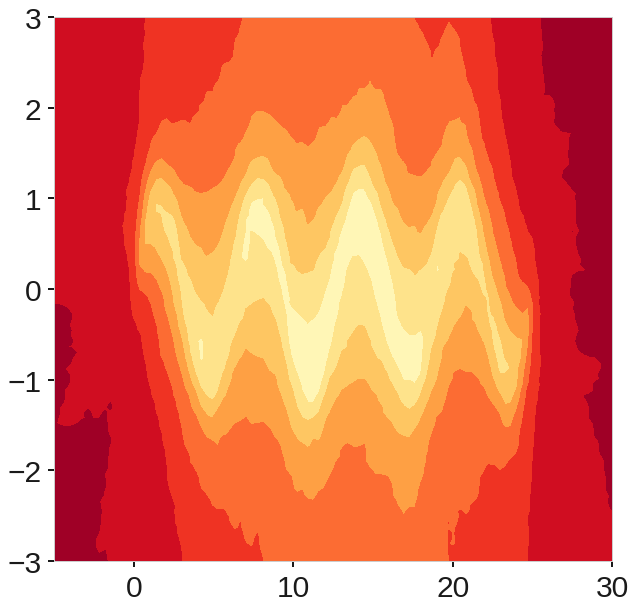}
		\caption{Extended IF}
		\label{fig:Sinusoid_ScoreMap_Extended_Result}
	\end{subfigure}
	\caption{Comparison of the standard Isolation Forest with rotated Isolation Forest, and Extended Isolation Forest for the case of sinusoid.}
	\label{fig:Sinusoid_ScoreMap_Result}
\end{figure}

Figure \ref{fig:Sinusoid_ScoreMap_Result} shows a comparison of the results for this case. The improvement in both cases over the Standard Isolation Forest is clear. The structure of the data is better preserved, and the anomaly score map is representative of the data. In the case of the Extended Isolation Forest, the score map tracks the sinusoidal data even more tightly than the rotated case.

\subsection{Variance of the anomaly scores}

Besides looking at anomaly score maps, we can consider some other metrics in order to compare the Extended Isolation Forest with the standard one. One such metric is looking at the mean and variance of the anomaly scores of points distributed along roughly constant score lines for various cases. The advantage of this metric is that we can test this for higher dimensional data.

\begin{figure}[H]
	\centering
	\begin{subfigure}{.15\textwidth}
		\centering
		\includegraphics[width=1\linewidth]{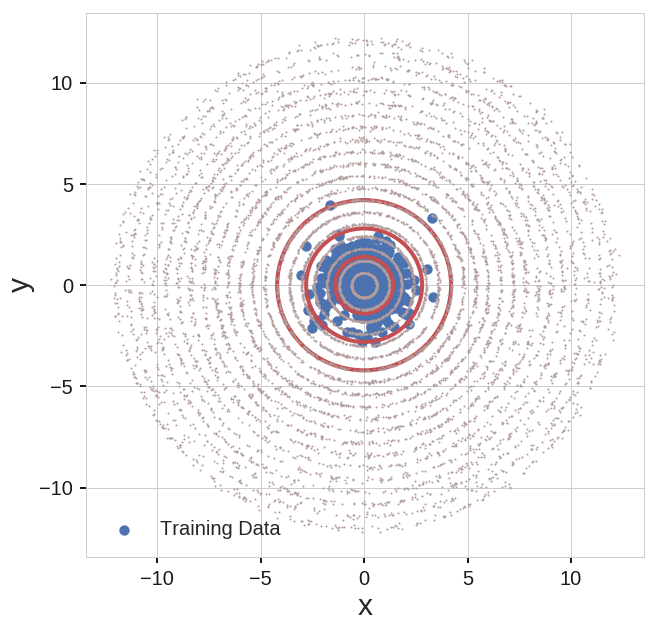}
		\caption{Data}
		\label{fig:Data_Radial_Grid_Blob}
	\end{subfigure}
	\begin{subfigure}{.15\textwidth}
		\centering
		\includegraphics[width=1\linewidth]{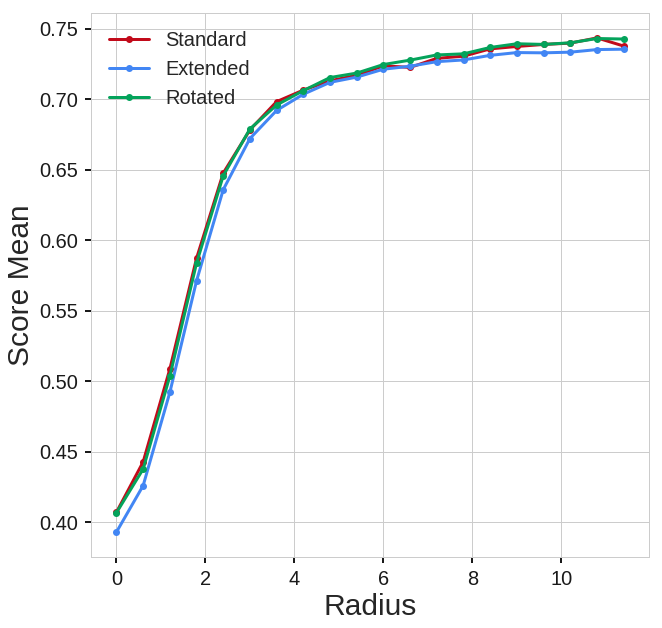}
		\caption{Score Mean}
		\label{fig:MeanvsRadius_2D_Blob}
	\end{subfigure}
	\begin{subfigure}{.15\textwidth}
		\centering
		\includegraphics[width=1\linewidth]{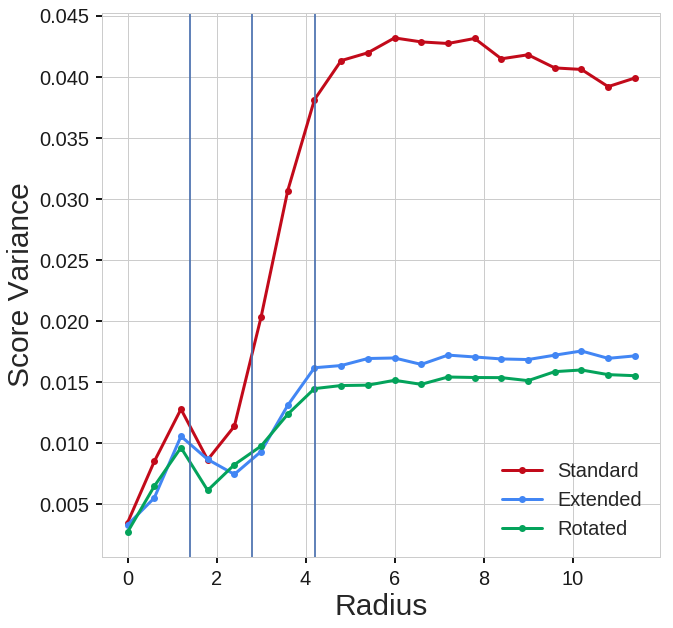}
		\caption{Score Variance}
		\label{fig:STDvsRadius_2D_Blob}
	\end{subfigure}
	\caption{Mean and variance of the scores for the data points located a concentric distance from the center.}
	\label{fig:ScoreVariance}
\end{figure}

Consider our randomly distributed data in two dimensions shown in Figure \ref{fig:Data_Radial_Grid_Blob}. The blue dots represent the training data, used to create the forest in each case. The gray dots form concentric circles. Since our data is normally distributed, anomaly scores along each circle should remain more or less a constant. The red lines represent the circles at $1\sigma$, $2\sigma$, and $3\sigma$ of the original normal distribution from which the training data is sampled.
We will run the data  from the gray circles through our trained Isolation Forest for the standard, the rotated and the extended cases, and look at the variation of the score as a function of distance from the center.

From Figure \ref{fig:MeanvsRadius_2D_Blob} we can see that the mean score varies the same way among the three methods. It increases quickly within $3\sigma$ from the center and then keeps monotonically increasing. At higher values the rate of increase slows down until it converges far away from the center where the blob resembles a single point.

As far as the variance of the scores goes, looking at Figures \ref{fig:Blob_ScoreMap_Result} and \ref{fig:STDvsRadius_2D_Blob}, we can see for very small anomaly scores (close to the center of the blob), the variance is quite small in all three cases. All three methods are able to detect the center of concentration of data quite well. This however, is not the goal for anomaly detection algorithms. After about $3\sigma$, the variance among the scores computed by the Extended Isolation Forest, and the case of Isolation Forest with rotated trees, settle to a much smaller value compared to the standard Isolation Forest, which translates to a much more robust measurement. It is notable that this inconsistency occurs in regions of high anomaly, which is very important for an anomaly detection algorithm. In short, for data in regions of high anomaly likelihood, the standard Isolation Forest produces anomaly scores which vary largely depending on alignment of these data points with the training data and the axes, rather than whether they are true anomalous points or not. This increases the chances of false positives in scoring stage.

\begin{figure}[H]
	\centering
	\begin{subfigure}{.15\textwidth}
		\centering
		\includegraphics[width=1\linewidth]{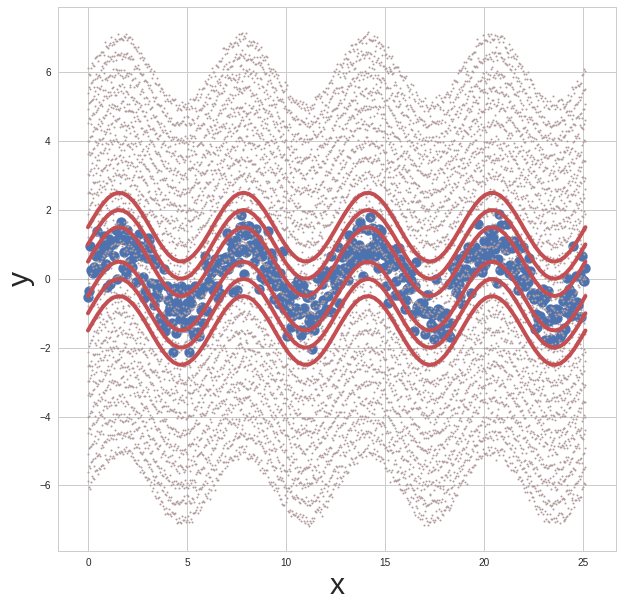}
		\caption{Data}
		\label{fig:Data_Sin_Grid}
	\end{subfigure}
	\begin{subfigure}{.15\textwidth}
		\centering
		\includegraphics[width=1\linewidth]{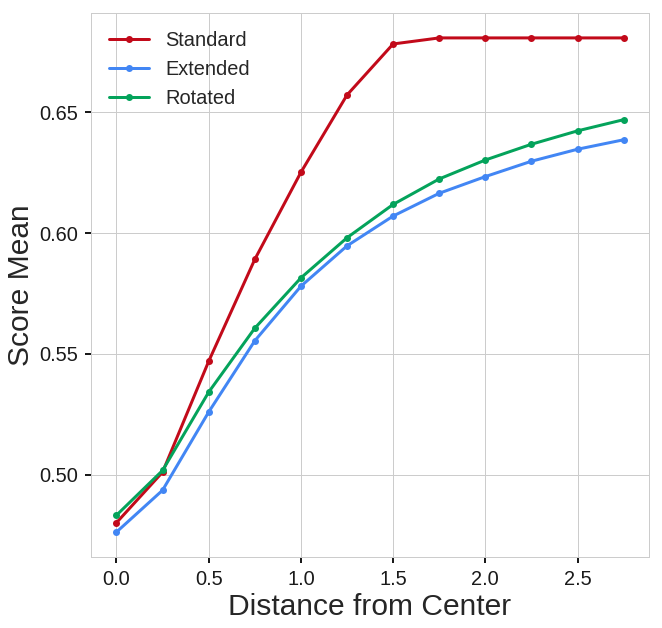}
		\caption{Score Mean}
		\label{fig:MeanvsRadius_Sin}
	\end{subfigure}
	\begin{subfigure}{.15\textwidth}
		\centering
		\includegraphics[width=1\linewidth]{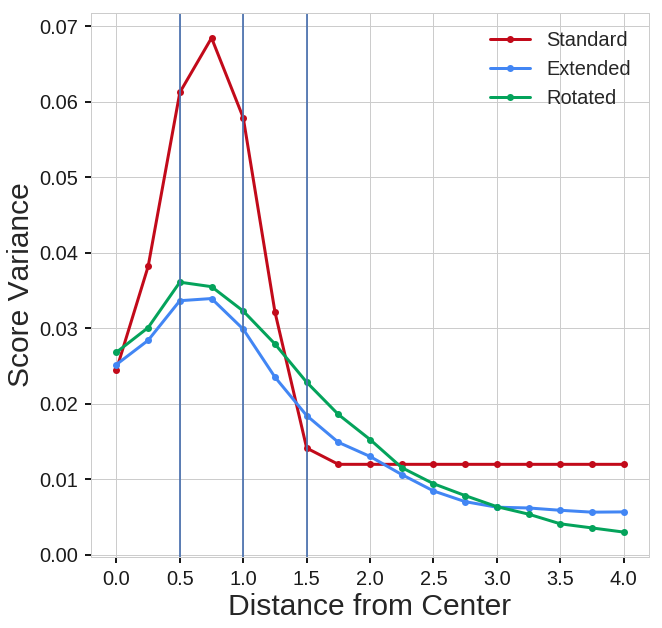}
		\caption{Score Variance}
		\label{fig:STDvsRadius_Sin}
	\end{subfigure}
	\caption{Mean and variance of the scores for data points at fixed distances from the center line.}
	\label{fig:ScoreVariance_with_data}
\end{figure}

We can do a similar study for the sine curve by considering lines at a similar distances above and below where the data points lie as shown in Figure \ref{fig:Data_Sin_Grid}. We assume data lying on each line to be scored more or less the same. This is however not quite true. If the data stretched to infinity in the horizontal direction, or we had periodic boundary conditions in that direction, this would indeed be true. However, since our data is bounded in the horizontal direction, there will be variations in the scores along each line as in the map shown in Figure \ref{fig:Sinusoid_ScoreMap_Extended_Result}.

Nonetheless we performed the analysis as shown in Figure \ref{fig:ScoreVariance_with_data}. Similarly to the previous plots, the three red lines on each side of the center line correspond to $1\sigma$, $2\sigma$, and $3\sigma$ from the Gaussian noise added to the sine curve. The gray lines represent data points that lie at constant distance from the center of the data. These are points that are used in the scoring stage. The blue data points are used for training. Figure \ref{fig:MeanvsRadius_Sin} shows the change in the mean score as we move out on both sides in the $y$ direction. We observe that in this case, the mean scores behave differently. The mean score for the standard Isolation forest reaches a constant value very quickly. This is consistent with what we observed in the score maps, figure \ref{fig:Sinusoid_ScoreMap_Generic_Result}. We saw that the standard Isolation Forest was unable to capture the structure present in this dataset, and treated it as a rectangular blob. The rapid rise of the anomaly score for this case represents moving through the boundary of this rectangular region into a high anomaly score area. Outside this region, all data points look the same to the standard Isolation Forest. This rapid rise is undesirable. In addition to not capturing the structure of the data, these scores are much more sensitive to choosing an anomaly threshold value which results in reducing AUCROC, as we will see below. In the other two cases, we observe a similar trend as before where the mean is converging to a steady score value, but much more slowly, making it less sensitive to choosing threshold values.

Figure \ref{fig:STDvsRadius_Sin} shows the variance of the scores obtained. We notice that, as before, the variance is much higher for the standard Isolation Forest, and lower for the extended Isolation Forest. After $3\sigma$ the variance for the standard case drops rapidly to a constant value but higher than that obtained by EIF, as the mean score reaches a steady value. The constant value here is a clear indication that the standard Isolation Forest is unable to capture any useful information outside the rectangular region discussed before. The EIF again provides a more robust scoring prediction.

In addition to the two examples above, we can perform the same analysis on higher dimensional data. We choose 3-D and 4-D blobs of normally distributed data in space around the origin with unit variance in all directions, similarly to the 2-D case. Recall that as soon as we move beyond the two dimensional case, we have more levels of extension that we can consider with regards to generating the plane cuts. These are represented by Ex$n$ as discussed above. We do not include the rotated case in the following analysis.

\begin{figure}[H]
	\centering
	\begin{subfigure}{.24\textwidth}
		\centering
		\includegraphics[width=0.9\linewidth]{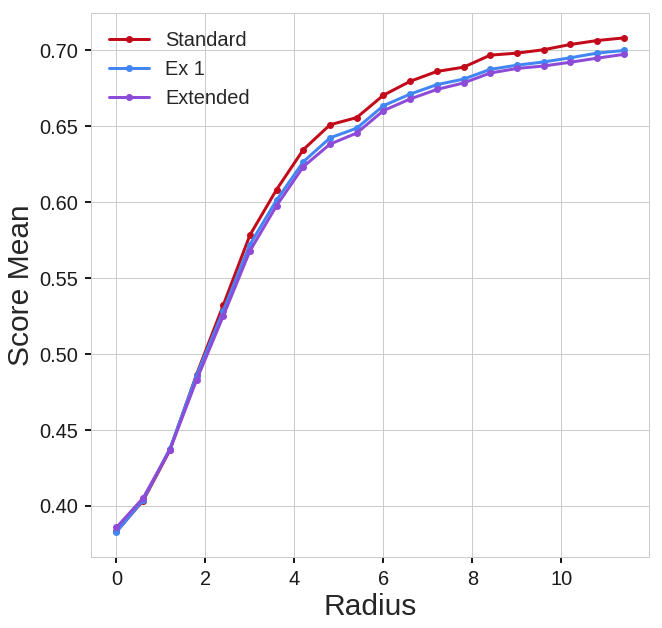}
		\caption{3-D Blob, mean of the scores}
		\label{fig:MeanvsRadius_3D_Blob}
	\end{subfigure}
	\begin{subfigure}{.24\textwidth}
		\centering
		\includegraphics[width=0.9\linewidth]{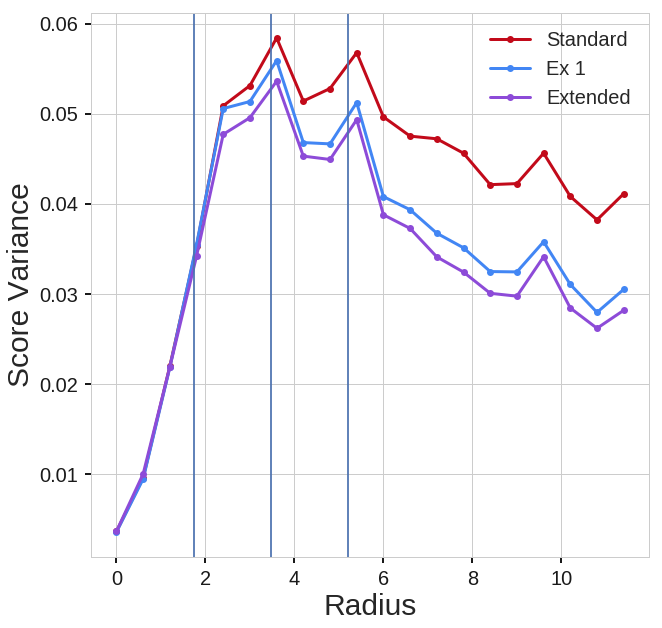}
		\caption{3-D Blob, variance of the scores}
		\label{fig:STDvsRadius_3D_Blob}
	\end{subfigure}
	\caption{Mean and variance for concentric points around a normally distributed 3-D blob}
\label{fig:STDvsRadius_3D}
\end{figure}

\begin{figure}[H]
	\begin{subfigure}{.24\textwidth}
		\centering
		\includegraphics[width=0.9\linewidth]{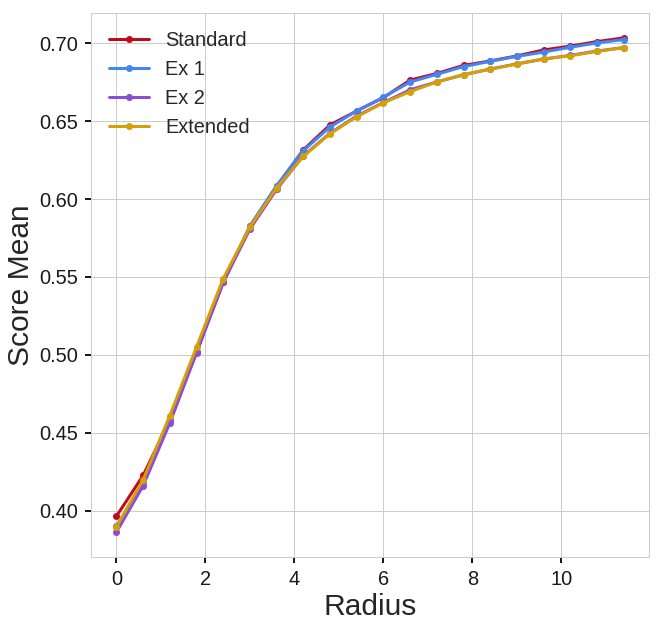}
		\caption{4-D Blob, mean of the scores}
		\label{fig:MeanvsRadius_4D_Blob}
	\end{subfigure}
	\begin{subfigure}{.24\textwidth}
		\centering
		\includegraphics[width=0.9\linewidth]{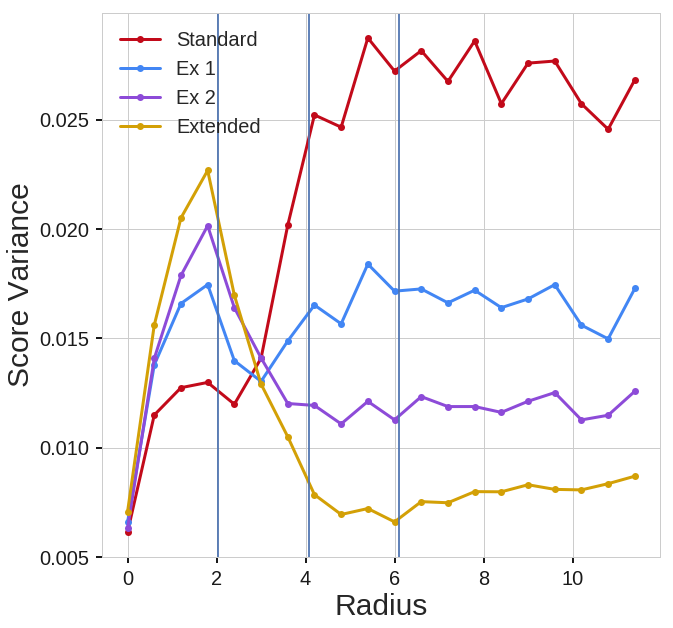}
		\caption{4-D Blob, variance of the scores}
		\label{fig:STDvsRadius_Blob}
	\end{subfigure}
	\caption{Mean and variance for concentric points around a normally distributed 4-D blob}
	\label{fig:STDvsRadius_4D}
\end{figure}

As before and not surprisingly, we can see from Figures \ref{fig:STDvsRadius_3D} and \ref{fig:STDvsRadius_4D} that the mean score increases with increasing radius very similarly in all cases, which was also observed in the 2-D blob case. However, again we see in the region beyond $3\sigma$, which is where the anomalies lie, the Extended Isolation Forest produces much lower values in the score variance. We also observe that for each case, the variance decreases as the extension level increases. In other words, the Extended Isolation Forest, i.e. the fully general case (when the planes subdividing the data have no constrains and the normal can point in any direction within all the dimensions of the problem), produces the most reliable and robust anomaly scores.

\subsection{AUC Comparison}

In this section we report AUC values for both ROC and PRC and compare the two methods. Initially, we produce these results for the examples provided earlier, since these examples are designed to highlight the specific shortcoming of Isolation Forest studied in this paper. We then report the AUC for ROC and PRC for some real world benchmark datasets of varying sizes and dimensions.

In order to obtain the AUC values for the example datasets, we first train both algorithms on 2000 data points in each case distributed as seen previously. We then add 200 anomalous points, score the data, and compute AUC's. The results are summarized in table \ref{tab:AUCSynth}.

\begin{table}[H]
	\centering
	\caption{AUC values for both ROC and PRC for example datasets using standard Isolation Forest and Extended Isolation Forest}
	\label{tab:AUCSynth}
	\begin{tabular}{ |c|cc|cc| }
		\hline
		        &  \multicolumn{2}{|c|}{AUC ROC} &  \multicolumn{2}{|c|}{AUC PRC}  \\
		Data & \scriptsize{iForest} & \scriptsize{EIF}    & \scriptsize{iForest} & \scriptsize{EIF}\\
 		\hline
		Single Blob & 0.919 & 0.999 &0.800  & 0.999\\ \hline
		Double Blob & 0.869 & 0.999 & 0.303   & 0.997 \\ \hline
		Sinusoid & 0.809 & 0.924 & 0.430   & 0.504\\ \hline
	\end{tabular}
\end{table}

In all cases the improvement is obvious, especially in the case of the double blob, where we have ``ghost'' regions, and the sinusoid, where we have a high variability in the structure of the dataset. This highlights the importance of understanding the problem that arises due to using axis-parallel hyperplanes.

We perform the same analysis on some real world benchmark datasets. Table \ref{tab:BenchmarkData} lists the datasets used as well as their properties. The results are summarized in table \ref{tab:AUCBenchmark}.

\begin{table}[H]
	\centering
	\caption{Table of data properties}
	\label{tab:BenchmarkData}
	\begin{tabular}{ |c|c|c|c| }
		\hline
		& Size &  Dimension & \% Anomally  \\ \hline
		Cardio & 1831 & 21 & 9.6 \\ \hline
		ForestCover & 286048 & 10 & 0.9 \\ \hline
		Ionosphere & 351 & 33 & 36 \\ \hline
		Mammography & 11183 & 6 & 2.32 \\ \hline
		Satellite & 6435 & 36 & 32 \\ \hline
	\end{tabular}
\end{table}

\begin{table}[H]
	\centering
	\caption{AUC values for both ROC and PRC for benchmark datasets using standard Isolation Forest and Extended Isolation Forest}
	\label{tab:AUCBenchmark}
	\begin{tabular}{ |c|cc|cc| }
		\hline
		&  \multicolumn{2}{|c|}{AUC ROC} &  \multicolumn{2}{|c|}{AUC PRC}  \\
		Data & \scriptsize{iForest} & \scriptsize{EIF}    & \scriptsize{iForest} & \scriptsize{EIF}\\
		\hline
		Cardio & 0.888 & 0.915 & 0.466  & 0.483\\ \hline
		ForestCover & 0.809 & 0.924 & 0.430   & 0.504\\ \hline
		Ionosphere & 0.85 & 0.913 & 0.877  & 0.893 \\ \hline
		Mammography & 0.859 & 0.862 & 0.4198 & 0.4271 \\ \hline
		Satellite & 0.714 & 0.778 & 0.783 & 0.808 \\ \hline
	\end{tabular}
\end{table}

We can see the improved AUC values in the case of Extended Isolation Forest. Even though in some cases the improvement may seem modest, in others, there is an appreciable difference. Naturally, the axis-parallel limitation has more effects on some datasets than other. Nevertheless, we can see that the EIF in general achieves better results than the standard Isolation Forest.

\subsection{Convergence of Anomaly Scores}

\begin{figure}[H]
	\centering
	\begin{subfigure}{.15\textwidth}
		\centering
		\includegraphics[width=1\linewidth]{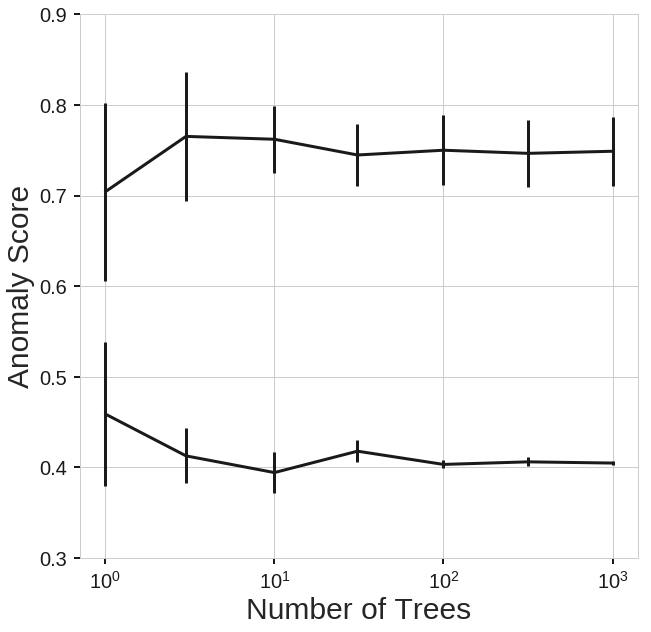}
		\caption{Standard Isolation Forest}
		\label{fig:MeanvsTrees_2D_Blob_Generic}
	\end{subfigure}
	\begin{subfigure}{.15\textwidth}
		\centering
		\includegraphics[width=1\linewidth]{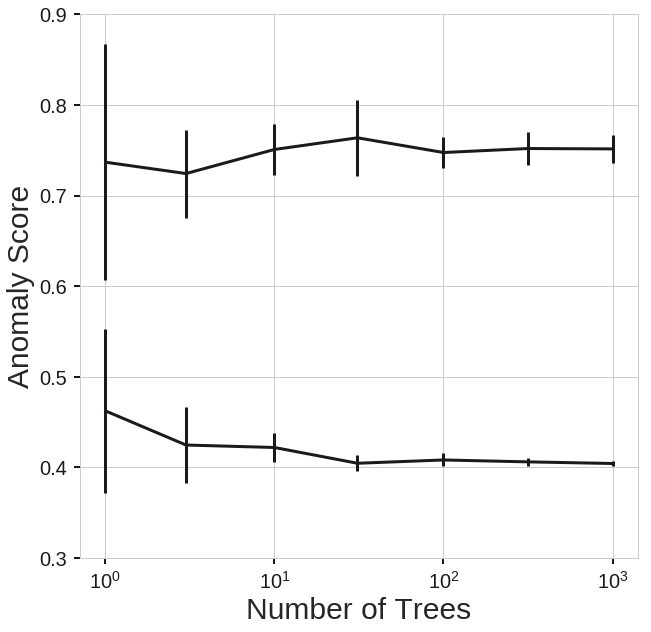}
		\caption{Rotated Isolation Forest}
		\label{fig:MeanvsTrees_2D_Blob_Rot}
	\end{subfigure}
	\begin{subfigure}{.15\textwidth}
		\centering
		\includegraphics[width=1\linewidth]{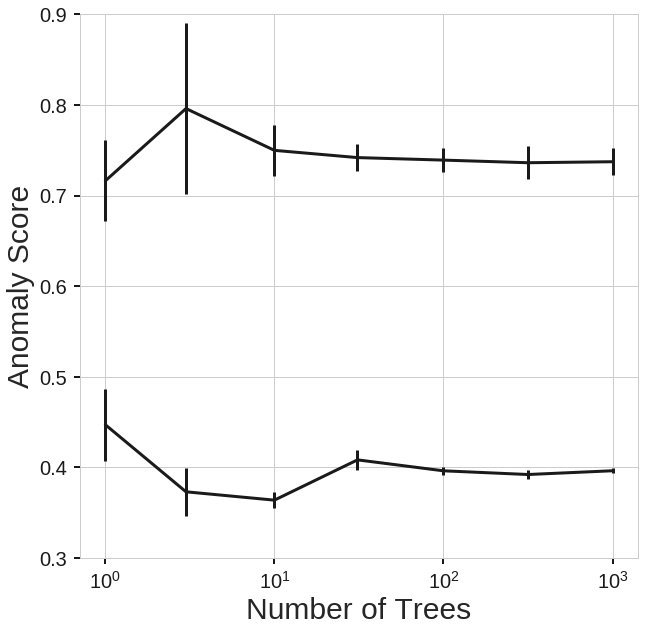}
		\caption{Extended Isolation Forest}
		\label{fig:MeanvsTrees_2D_Blob_Gen1}
	\end{subfigure}
	\caption{Convergence plots for the 2-D Blob case and the three methods described in the text.}
	\label{fig:MeanvsTrees_2D_Blob}
\end{figure}

In a final comparison, we would like to find out how efficient the Extended Isolation Forest is compared to the standard case. For this purpose we consider convergence plots for the anomaly score as a function of forest size (the number of trees). We perform this for the 2-D, 3-D, and 4-D blobs. In the case of 2-D we also include the method of tree rotations, described before. Figure \ref{fig:MeanvsTrees_2D_Blob} shows the convergence plots for the standard, rotated and Extended Isolation forest.

\begin{figure}[H]
	\centering
	\begin{subfigure}{.24\textwidth}
		\centering
		\includegraphics[width=1\linewidth]{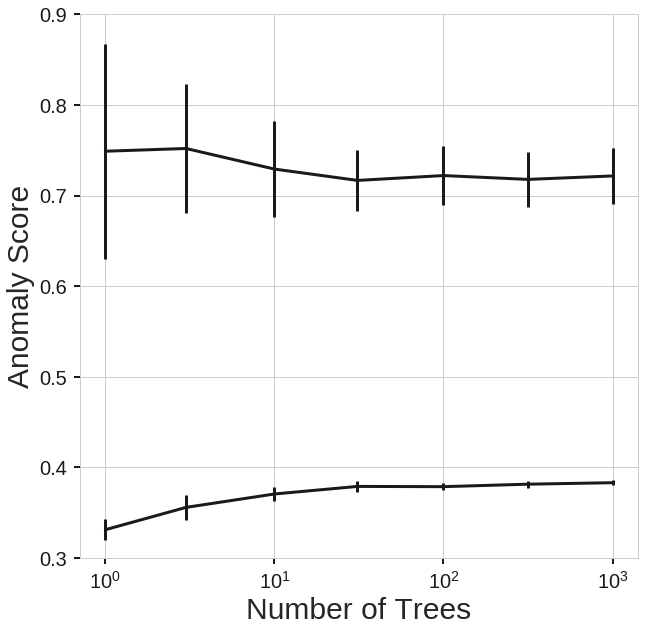}
		\caption{Standard Isolation Forest}
		\label{fig:MeanvsTrees_3D_Blob_Generic}
	\end{subfigure}
	\begin{subfigure}{.24\textwidth}
		\centering
		\includegraphics[width=1\linewidth]{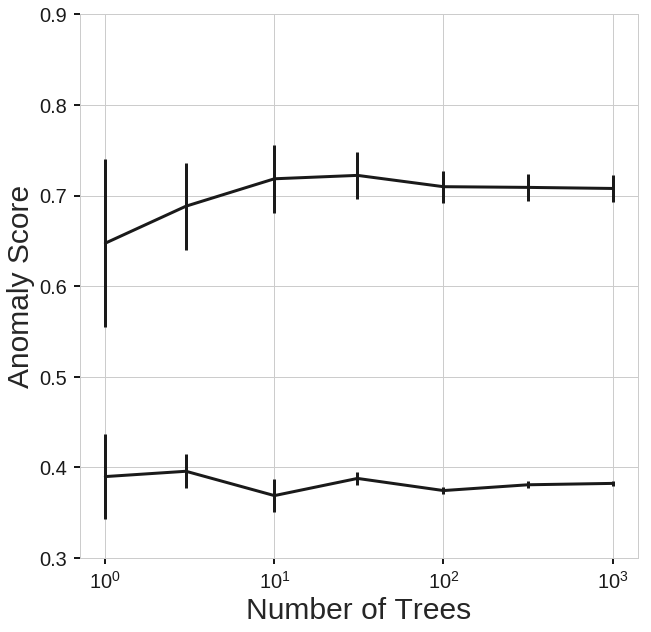}
		\caption{Extended Isolation forest}
		\label{fig:MeanvsTrees_3D_Blob_Gen2}
	\end{subfigure}
	\caption{Convergence plots for the 3-D blob for the standard and fully extended cases.}
	\label{fig:MeanvsTrees_3D_Blob}
\end{figure}

\begin{figure}[H]
	\centering
	\begin{subfigure}{.24\textwidth}
		\centering
		\includegraphics[width=1\linewidth]{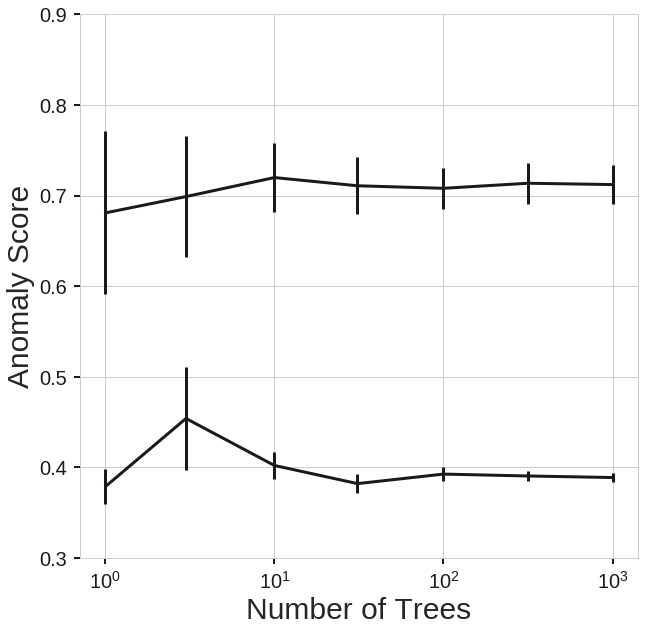}
		\caption{Standard Isolation Forest}
		\label{fig:MeanvsTrees_4D_Blob_Generic}
	\end{subfigure}
	\begin{subfigure}{.24\textwidth}
		\centering
		\includegraphics[width=1\linewidth]{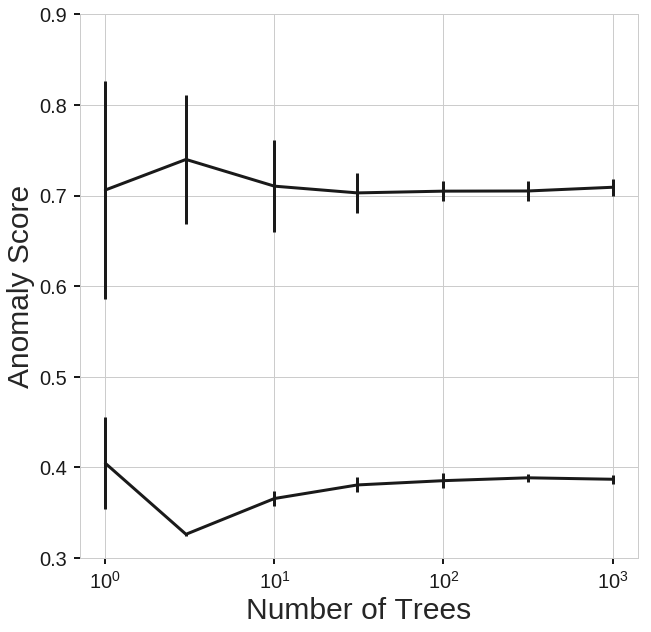}
		\caption{Extended Isolation forest}
		\label{fig:MeanvsTrees_4D_Blob_Gen3}
	\end{subfigure}
	\caption{Convergence plots for the 4-D blob for the standard and fully extended cases.}
	\label{fig:MeanvsTrees_4D_Blob}
\end{figure}

Figures \ref{fig:MeanvsTrees_3D_Blob} and \ref{fig:MeanvsTrees_4D_Blob} show the same plots for the 3-D blob and 4-D blobs respectively, except for the Rotated Isolation Forest case.

For the convergence plots, the error bars show the variance among the scores of 500 data points tested along constant level sets as before (concentric shells of 2 and 3-spheres). As we can see, in all the cases the score settles very quickly as a function of the forest size. There might be a slight difference between the nominal and anomalous points but the differences are very small. Moreover we find no appreciable difference between the standard Isolation Forest and the Extended Isolation Forest, including all the extension levels. In the case of standard Isolation Forest, the variances are much higher, consistent with the results found in the previous subsection, but in terms of rates of convergence, there is not much difference which indicates that the number of trees on each method should be the same.

We have seen that the Extended Isolation Forest is able to improve on the robustness of the scores generated compared to the standard case of Isolation Forest. It accomplishes this without sacrificing computational efficiency as evidenced by the convergence plots presented in this section. The fully extended case seems to produce the best results when higher dimensional data are considered.

\section{Conclusions} \label{sec:Conclusion}
We have presented an extension to the anomaly detection algorithm known as Isolation Forest \cite{liu2008isolation}. The motivation for the study arose from the observation that score maps in two dimensional datasets demonstrated unexpected artifacts. We showed these artifacts can be traced to the branching procedure in the algorithm itself. The branching procedure followed slicing data along random values of randomly selected features. This introduced a bias in terms of the number of branching operations performed based on the location of the data point with respect to the coordinate frame. Since the length of tree branches are used directly in computing the anomaly scores, we saw regions in the domain that showed inconsistent anomaly scores, introducing artificial zones of higher/lower scores which are not present in the original data.

In order to remedy the situation we proposed two different approaches. First we showed the score maps can be improved if the data undergoes a rotation before the construction of each tree. The second approach, Extended Isolation Forest, allows the branching hyperplanes to take on any slope as opposed to hyperplanes only parallel to the coordinate frame. This extension in the algorithm completely resolves the bias introduced in the case of standard Isolation Forest. We show the algorithm is readily extended to high dimensional datasets, while it possesses $Ex=N-1$ levels of extensions for an $N$ dimensional dataset, with $Ex=0$ being identical to the standard Isolation Forest. All of these extension levels show better performance than the standard case, and we suggest the EIF should be preferred over the rotation approach.

We showed score maps for various examples. We also showed that for the region of anomaly, the standard Isolation Forest algorithm produces high variance in the scores along constant level sets of anomaly scores, while the Extended Isolation Forest resulted in remarkably smaller variances which decreased as the extension level increased. We presented AUC for ROC and PRC for all the examples considered in the paper as well as various real world benchmark data. The EIF performed consistently better then the standard Isolation Forest in all cases considered.

The results of this extension are more reliable and robust anomaly scores, and in some cases more accurate detection of structure of a given dataset. We additionally saw these results can be achieved without sacrificing computational efficiency.   We have made a Python implementation of the Extended Isolation Forest algorithm publicly available \footnote{https://github.com/sahandha/eif}, accompanied by example Jupyter notebooks to reproduce the figures in this paper.

\section*{Acknowledgments}

We would like to thanks Seng Keat Yeoh for useful comments and discussion in regards implementation.
MCK work has been supported by grant projects NSF AST 07-15036 and NSF AST 08-13543.

\bibliographystyle{plain}
\bibliography{./BibFiles/bib}

\end{document}